\documentclass[10pt,journal,compsoc]{IEEEtran}

\usepackage{booktabs}
\usepackage{multirow}
\usepackage{amssymb}
\usepackage{amsmath}
\usepackage{graphicx}
\usepackage{bm}
\usepackage{bbm}
\usepackage{amsthm}

\usepackage{array}
\usepackage{color}
\usepackage{bm}
\usepackage{subfigure}
\usepackage{float}
\usepackage{subfig}
\usepackage[linesnumbered,algoruled,boxed,lined,noend]{algorithm2e}
\usepackage{caption}
\allowdisplaybreaks
\begin{document}

\title{Aerial Reliable Collaborative Communications for Terrestrial Mobile Users via Evolutionary Multi-Objective Deep Reinforcement Learning}

\author{Geng~Sun,~\IEEEmembership{Senior Member,~IEEE,}
Jian~Xiao,
Jiahui~Li,
Jiacheng~Wang,
Jiawen~Kang,~\IEEEmembership{Senior Member,~IEEE,}
Dusit~Niyato,~\IEEEmembership{Fellow,~IEEE,}
Shiwen~Mao,~\IEEEmembership{Fellow,~IEEE}
  
\thanks{This work is supported in part by the National Natural Science Foundation of China (62272194, 62172186), in part by the Science and Technology Development Plan Project of Jilin Province (20230201087GX), in part by the Postdoctoral Fellowship Program of CPSF (GZC20240592), in part by China Postdoctoral Science Foundation General Fund (2024M761123), in part by the Scientific Research Project of Jilin Provincial Department of Education (JJKH20250117KJ), in part by the National Research Foundation, Singapore, in part by Infocomm Media Development Authority under its Future Communications Research \& Development Programme, in part by Defence Science Organisation (DSO) National Laboratories under the AI Singapore Programme (FCP-NTU-RG-2022-010 and FCP-ASTAR-TG-2022-003), in part by Singapore Ministry of Education (MOE) Tier 1 (RG87/22 and RG24/24), in part by the NTU Centre for Computational Technologies in Finance (NTU-CCTF), in part by the RIE2025 Industry Alignment Fund - Industry Collaboration Projects (IAF-ICP) (Award I2301E0026), administered by A*STAR, and in part by Alibaba Group and NTU Singapore through Alibaba-NTU Global e-Sustainability CorpLab (ANGEL). (Corresponding author: Jiahui Li.)
\par Geng Sun is with the College of Computer Science and Technology, Jilin University, Changchun 130012, China, and with Key Laboratory of Symbolic Computation and Knowledge Engineering of Ministry of Education, Jilin University, Changchun 130012, China; he is also affiliated with the College of Computing and Data Science, Nanyang Technological University, Singapore 639798 (e-mail: sungeng@jlu.edu.cn).
\par Jian Xiao and Jiahui Li are with the College of Computer Science and Technology, Jilin University, Changchun 130012, China (e-mails: dajianer@foxmail.com; lijiahui@jlu.edu.cn).
\par Jiacheng Wang and Dusit Niyato are with the School of Computer Science and Engineering, Nanyang Technological University, Singapore 639798 (e-mail: jcwang\_cq@foxmail.com; dniyato@ntu.edu.sg).
\par Jiawen Kang is with the School of Automation, Guangdong University of Technology, Guangzhou 510641, China (e-mail: kavinkang@gdut.edu.cn).
\par Shiwen Mao is with the Department of Electrical and Computer Engineering, Auburn University, Auburn, AL 36849-5201 USA (e-mail: smao@ieee.org).
}
}

\IEEEtitleabstractindextext{
%
%
\begin{abstract}

\par Unmanned aerial vehicles (UAVs) have emerged as the potential aerial base stations (BSs) to improve terrestrial communications. However, the limited onboard energy and antenna power of a UAV restrict its communication range and transmission capability. To address these limitations, this work employs collaborative beamforming through a UAV-enabled virtual antenna array to improve transmission performance from the UAV to terrestrial mobile users, under interference from non-associated BSs and dynamic channel conditions. Specifically, we introduce a memory-based random walk model to more accurately depict the mobility patterns of terrestrial mobile users. Following this, we formulate a multi-objective optimization problem (MOP) focused on maximizing the transmission rate while minimizing the flight energy consumption of the UAV swarm. Given the NP-hard nature of the formulated MOP and the highly dynamic environment, we transform this problem into a multi-objective Markov decision process and propose an improved evolutionary multi-objective reinforcement learning algorithm. Specifically, this algorithm introduces an evolutionary learning approach to obtain the approximate Pareto set for the formulated MOP. Moreover, the algorithm incorporates a long short-term memory network and hyper-sphere-based task selection method to discern the movement patterns of terrestrial mobile users and improve the diversity of the obtained Pareto set. Simulation results demonstrate that the proposed method effectively generates a diverse range of non-dominated policies and outperforms existing methods. Additional simulations demonstrate the scalability and robustness of the proposed CB-based method under different system parameters and various unexpected circumstances.

\end{abstract}

%
%
\begin{IEEEkeywords}
UAV communications, collaborative beamforming, random mobility models, multi-objective optimization, and multi-objective reinforcement learning.
\end{IEEEkeywords}
}
\maketitle
%
%
\section{Introduction}
\label{sec:Introduction}

\IEEEPARstart{A}{s} a result of manufacturing improvements and cost reductions, unmanned aerial vehicles (UAVs) play an essential role in various domains, such as academia, industry, and military \cite{Sun2024}. Owing to their high relocation flexibility and excellent maneuverability, UAVs are increasingly expected to function as aerial base stations (BSs) or relays to enhance terrestrial communications by facilitating data access from the sky \cite{liang2024multiobjective,Zhang2024}. Moreover, UAVs can be rapidly deployed in disaster areas where terrestrial BSs are absent, which can provide crucial emergency communication services. However, a single UAV can only serve a limited area due to the limitations on the onboard energy and transmit power \cite{zeng2019accessing}, and these constraints prevent the UAVs from meeting the communication needs of remote users. Moreover, time-varying channels and interference from other BSs complicate achieving satisfactory communication rates. Thus, enhancing the transmission capabilities of UAVs is essential for providing reliable, high-rate, and extensive coverage communication services.

\par Collaborative beamforming (CB) has been demonstrated to be an effective method for enhancing both the signal strength and directivity without requiring alterations to existing devices. Consequently, we introduce CB to improve the communication performance of UAVs~\cite{jayaprakasam2017}. Specifically, multiple UAVs can collaborate to form a UAV-enabled virtual antenna array (UVAA), and within this framework, the UVAA elements synchronize and adjust their carrier phases to generate a high-gain mainlobe directed toward the remote user. As such, CB has the ability to amplify the received power at the destination by a factor proportional to the square of the number of UAV elements, thereby significantly boosting the transmission efficiency of a UAV swarm, which extends the communication distances and improves the interference resistance. 

\par The performance of such UVAA systems is subject to a broader range of variables than traditional CB applications in terrestrial networks. Notably, the stochastic spatial distribution of UAVs within a three-dimensional (3D) domain can disrupt the beam pattern of the UVAA. To mitigate this and achieve higher gains, UAVs can move to more favorable positions, which may optimize the beam pattern but at the cost of additional energy for UAV flights. Furthermore, the excitation current weight assigned to each UAV is crucial, affecting both the directivity and the transmission rate of the UVAA. Thus, it is important to design an efficient method to determine the excitation current weights and positions of the UAVs to enhance transmission rates while conserving energy during UAV movement. In addition, with the rapid increase in mobile equipment, the mobility of user devices has become a crucial factor that affects the communication quality \cite{Nikooroo2022optimal}, \cite{Yoon2020rendezvous, Sun2023, 10607924}. Existing works primarily focus on CB communication frameworks for serving static terrestrial devices \cite{Li2023multi-objective, Zhang2024, Sun2024, sun2022secure, Xu2023}, which may overlook the potential mobility of terrestrial terminals in highly dynamic environments, such as pedestrians, robots, and smart vehicles. This dynamic necessitates that the UVAA system makes real-time control decisions without advance knowledge of the future locations of the users or environmental conditions. 

\par However, it is challenging to overcome these issues and fill in this gap. On the one hand, finding a trade-off between the transmission rate from the UVAA to the user and the energy consumption of the UAVs is difficult. This is because designing trajectories with high transmission rates to a terrestrial mobile user in multiple time slots requires the UAV to move frequently, potentially leading to an excessive increase in the additional energy consumption of UAVs. On the other hand, conventional offline optimization methods may be ineffective in highly dynamic environments with time-varying channels and user movement. Therefore, we aim to introduce a multi-objective online optimization method to address these challenges. Accordingly, the main contributions of this work can be detailed as follows:

\begin{itemize}
    \item \textit{CB-enabled UAV Reliable Mobile Communications System:} Aiming at the challenge of providing reliable communication to mobile ground users, we model a representative CB-enabled UAV reliable mobile communications system. Specifically, multiple UAVs form a UVAA to transmit information to a terrestrial mobile user, while contending with interference from non-associated BSs and time-varying channel conditions. In this system, we adopt realistic random mobility models with memory, \textit{i.e.,} the Gaussian-Markov model, to simulate the movement patterns of terrestrial mobile users. This allows us to create a system that is more aligned with real-world applications and addresses the challenges posed by user mobility on CB systems.
    
    \item \textit{Long-term Multi-objective Optimization Problem Formulation:} In the considered system, the transmission rates and energy consumption of the UAVs in UVAA have inherent trade-offs over time. In this case, improving the achievable rate from the UVAA to the user and reducing the energy consumption of UAVs simultaneously is challenging. Thus, we formulate a long-term multi-objective optimization problem (MOP) with the dual objectives of maximizing the total achievable rate and minimizing the overall flight energy consumption of UAVs. This problem explicitly considers the sequential and interdependent nature of UAV decisions across time slots. Thus, the long-term nature of the problem necessitates balancing short-term and sustainable performance gains over the entire operational period, making the problem non-trivial.
    
    \item \textit{Improved Multi-Objective Deep Reinforcement Learning Approach:} Given the NP-hard nature of the formulated MOP and challenges presented by time-varying environments, we propose an evolutionary multi-objective optimization proximal policy optimization with vectorized value function, long short-term memory (LSTM) networks, and hyper-sphere-based task selection (EMOPPO-VLH). Specifically, the algorithm extends the value function from the single-objective proximal policy optimization (PPO) algorithm into a vectorized form, so that enabling it to handle multiple optimization objectives simultaneously. Moreover, the algorithm integrates the LSTM networks to capture both short-term (\textit{e.g.}, multi-path fading) and long-term (\textit{e.g.}, user mobility) dependencies, thereby enhancing its capability to handle dynamic environments. In addition, we introduce a hyper-sphere-based task selection method to improve the diversity of the obtained Pareto set, thereby achieving a well-distributed set of solutions across the Pareto front. These innovations are tailored to address the challenges of the dynamic and multi-objective nature of the formulated problem in the designed system.
    
    \item \textit{Performance Evaluations and Analyses:} We validate the effectiveness of our proposed MOPPO-PLE algorithm through extensive numerical simulations. The results demonstrate its ability to generate a set of high-quality non-dominated policies, showing superior performance compared with other benchmark algorithms across different scales. Moreover, we analyze the results by using several evaluation metrics, including the inverted generational distance and hypervolume, providing insights into the effectiveness of our approach. It is also found that the mobility of users will not affect the effectiveness of the proposed method.
\end{itemize}

\par The remainder of this paper is organized as follows. Section~\ref{sec:Related Work} introduces some related works. Section~\ref{sec:Models and Preliminaries} details the models and preliminaries. The MOP is formulated in Section~\ref{sec: problem formulation and analysis}. The proposed EMOPPO-VLH is introduced in Section~\ref{sec:algorithm}. Section~\ref{sec:simulation results} provides the simulation results. Section~\ref{sec:discussion} presents some discussions. Finally, Section~\ref{sec:conclusion} concludes this paper.

%
%
\section{Related Work}
\label{sec:Related Work}

\par This section comprehensively introduces major related studies about UAV communications, collaborative beamforming, and multi-objective optimization.

\subsection{UAV-assisted Terrestrial Communications}
\label{subsec:UAV-assisted Terrestrial Communications}

\par Several existing works have utilized UAVs to support terrestrial communications. For instance, the authors in~\cite{xu2021robust} investigated a UAV-assisted device-to-device communication network wherein a UAV acts as an aerial BS to provide communication services for ground terminals. In this scenario, energy efficiency is optimized through the joint optimization of radio resource allocation and flight altitude, considering the imperfections in channel state and coordinate information. The authors in~\cite{Gao2022coverage} envisioned a search and rescue scenario in which a swarm of UAVs is employed to provide downlink communication coverage over an unknown mission area. The objective is to maximize the wireless coverage provided by the UAVs by designing the quasistationary deployment of the UAVs. The authors in~\cite{Lee2020aUAV-Mounted} studied an air-to-ground free space optical communication system, which provides communication between a UAV and terrestrial terminals, proposing a low-complexity approach aimed at maximizing the flight time of the UAV. The authors in~\cite{Xu2020multiuser} considered a robust resource allocation method for a multiuser downlink UAV communication system with the objective of minimizing total power consumption. Moreover, the authors in~\cite{Yan2019agametheory} investigated a UAV-assisted Internet-of-Things (IoT) communication network. In this study, a group of UAVs was dispatched in an urban area by using the wireless resources of the base station (BS) to serve IoT applications and proposed a game theory approach aimed at maximizing the communication rate for the users involved.

\par The distinctions between our study and the previously mentioned works can be analyzed as follows. First, prior research has not considered the use of CB as an alternative to the independent operation of UAVs in communication networks. This work introduces CB to extend communication ranges, enhance signal quality, and reduce the energy consumption of UAVs, especially in long-distance transmission scenarios affected by interference. Moreover, most of the aforementioned works focus on optimizing a single objective, such as coverage or communication rate. In real-world UAV-assisted communication systems, multiple conflicting objectives must often be considered, such as balancing the achievable rate with energy efficiency. In contrast, this work addresses this gap by incorporating a multi-objective optimization approach that balances both transmission performance and energy consumption, making it more suitable for practical deployments in dynamic environments.

\subsection{Collaborative Beamforming Methods}
\label{subsec:Collaborative Beamforming}

\par Several studies have explored the application of CB to enhance the transmission capabilities in various wireless communication scenarios. For example, the authors in~\cite{Bao2019astochastic} applied CB in wireless sensor networks and introduced a reinforcement learning approach aimed at optimizing the signal-to-noise ratio. However, such studies were limited to static sensor nodes. Notably, the application of CB in mobile nodes introduces greater complexity than in static environments. Moreover, the authors in  \cite{Mozaffari2019communicationsandcontrol} utilized UAVs to form a linear antenna array to enhance wireless communications, thereby minimizing the airborne service time by optimizing UAV locations and rotor speeds. However, this study was limited by its reliance on a simplified line-of-sight (LoS) channel model, which may not be applicable to realistic air-to-ground (A2G) communication links because of the lack of consideration for multipath fading. The authors in~\cite{Jung2022securityenergy} examined a secure communication network for multiple UAVs and explored a stochastic virtual antenna array to maximize energy efficiency. Moreover, the authors in~\cite{sun2022secure} investigated a novel UAV-assisted aerial relay system in which UAVs form a virtual antenna array to communicate with distant ground users by using CB, and they proposed a multi-objective optimization approach aimed at maximizing the secrecy rate and minimizing energy costs. 

\par While these studies demonstrate the efficiency and advantages of CB, none of them have explicitly explored its integration with dynamic terrestrial mobile users in more realistic CB-enabled communication environments. Such a gap is particularly challenging due to the uncertainty and rapid changes in such environments, requiring systems to exhibit high adaptability. In contrast, this work models and captures the mobility of the users, and proposes a DRL-based optimization method with real-time response ability, which can provide a more comprehensive and practical solution to the challenges of UAV-assisted communication networks in real-world settings.

\subsection{Multi-objective Optimization}
\label{Multi-objective Optimization}

\par In practical UAV-assisted communication networks, multiple conflicting optimization objectives often arise. Methods to address MOPs can generally be categorized into two main approaches which are traditional methods and deep reinforcement learning (DRL) techniques.

\par There have been several studies employing traditional methods to address the MOP within the context of UAV-assisted communication networks. For instance, the authors in~\cite{Liu2021resource} considered a multi-objective resource management optimization problem in heterogeneous cellular networks and designed a gravitational search algorithm aimed at minimizing the dispersion degree of throughputs and the total energy consumption. In~\cite{Muhammad2021performance}, a weighted Tchebycheff approach was introduced to concurrently maximize the achievable sum rate and minimizing the downlink transmission power in a UAV-assisted wireless communication network. The authors in~\cite{Shafique2020end-to-end} introduce a difference of convex optimization algorithm designed to minimize both the energy consumption and the SNR outage in a UAV-assisted data ferrying network. Moreover, the authors in~\cite{Li2023multi-objective} introduced and modified the multi-objective salp swarm algorithm for a UAV-assisted data harvesting and dissemination system, with the goals of reducing data transmission time, conserving energy for the UAV swarm, and enhancing secure performance. However, these traditional methods become impractical for real-time decisions in highly dynamic environments.

\par Some existing research considers the use of the DRL approach to solve the MOP. For example, the authors in~\cite{Sun2021aoi-energy-aware} considered a UAV-enabled IoT network, utilizing twin-delayed deep deterministic policy gradients to minimize the weighted sum of the age of information and UAV energy consumption. The authors in~\cite{Cui2020multi-agent} introduce a resource allocation algorithm based on multi-agent Q-Learning, aimed at optimizing both the achieved throughput and the power consumption. In \cite{Fu2021energy-efficient}, Q-learning was utilized to maximize the uplink throughput while minimizing the energy consumption in a UAV-based data collection system. The authors in~\cite{Zeng2021simultaneous} addressed a UAV trajectory optimization problem with the objectives of minimizing mission completion time and expected communication outage duration and introduced a dueling double deep Q network for UAV trajectory control. Moreover, the authors in~\cite{Zhao2022multi-agent} used a multi-agent deep deterministic policy gradient algorithm to maximize geographical fairness and minimize energy consumption in UAV trajectory control. 

\par Nevertheless, the aforementioned DRL methods employ single-policy approaches. For example, multiple objectives are usually combined into one reward by using various arithmetic methods, which complicates the determination of weights to harmonize these objectives effectively. Moreover, this method tends to yield one single solution, which may potentially overlook conflicts between objectives and reduce the solution space. Our previous work has proposed an evolutionary multi-objective DRL algorithm that may overcome this issue~\cite{10679228}. However, the previous work targeted periodic low Earth orbit (LEO) satellites, which cannot handle the high dynamics and temporal dependencies of the considered scenario raised by the mobile user. Moreover, the potential decision variables of this work such as UAV trajectories are continuous and high dimensions. Compared to our previous work that focused on static scenarios with fixed ground terminals and periodic LEO satellites, this work involves fundamentally different system components including 3D-controllable UAVs, randomly moving users, and highly dynamic channel conditions. Furthermore, from the mathematical perspective, it requires handling continuous decision variables, temporal dependencies, and more complex multi-objective optimization. As such, the considered scenario requires a significantly different algorithm design and improvement from our previous work in ~\cite{10679228}. Consequently, our objective is to introduce a novel evolutionary multi-objective DRL algorithm capable of deriving a collection of high-quality, non-dominated policies and handling high dynamics and temporal dependencies.

%
%
\section{System Models and Preliminaries}
\label{sec:Models and Preliminaries}

\par In this section, we first present the architecture of our proposed UAV-enabled A2G communication system. Then, we describe the system models, including the communication model, UAV movement model, and a memory-based user random mobility model. For ease of reference, a comprehensive list of the primary notations utilized throughout this study is provided in Table~\ref{tab:notation}.

%
%
\subsection{Network Segments}
\label{subsec:Network Segments}

\par
\begin{figure}[t]
	\centerline{\includegraphics[width=3.5in]{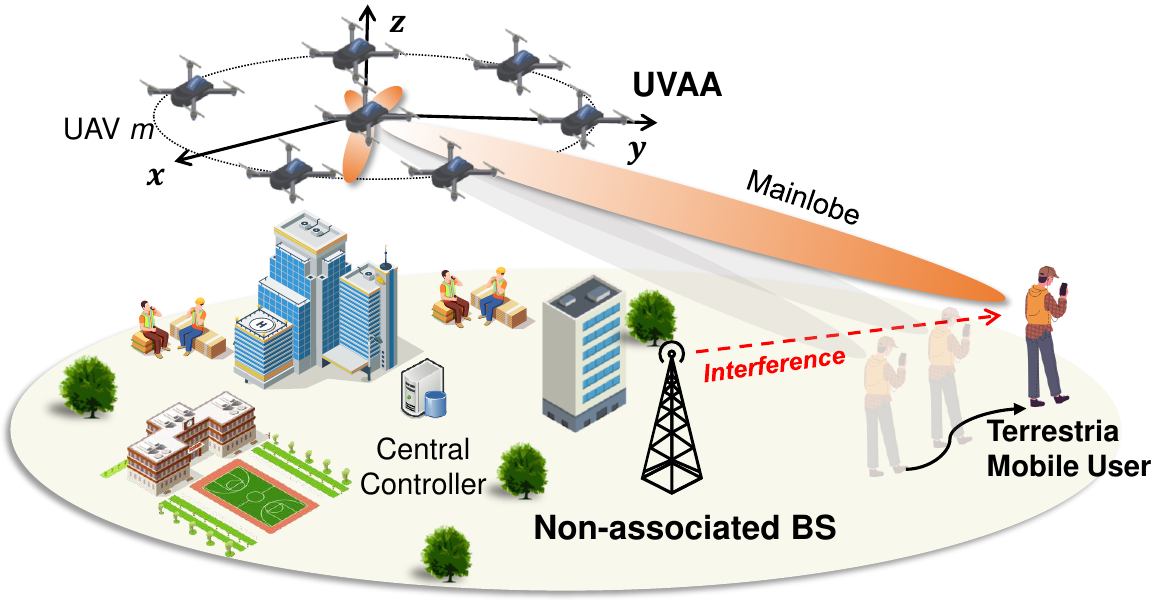}}
	\caption{A UAV-enabled A2G communication system, where a UAV swarm is deployed to transmit data to a remote terrestrial mobile user via CB. Moreover, this system has a central controller for controlling the UAVs and exists non-associated BS which may interfere with the communications.} 
	\label{fig.Sketch map}
\end{figure}

 As shown in Fig.~\ref{fig.Sketch map}, we consider a UAV-enabled A2G communication system which comprises the following elements:

\begin{itemize}

    \item A swarm of rotary-wing UAVs represented as $\mathcal{N} = \{1, 2, \ldots, N\}$, and these UAVs are used to transmit data or services to a terrestrial mobile user. Owing to their limited transmit power and complex network environment, a single UAV cannot establish a stable wireless link with the terrestrial mobile user.

    \item A terrestrial mobile user whose trajectory may change randomly over time, and it may move randomly within a fixed area.

    \item A non-association BS that may interfere with communication between the UAVs and the terrestrial mobile user. 

    \item A central controller is used to manage the UAVs and to execute computational tasks. We consider that the communications between UAVs and the central controller operate on their control channel, thereby ensuring no interference with the communications between the UAVs and the user.

\end{itemize}

\par To improve the communication range and stability of the UAV swarm while simultaneously mitigating the influence of BS interference, the system is designed to operate as follows. The UAVs will form a UVAA to enhance their transmission gain, thereby establishing a stable link to the remote terrestrial mobile user. Note that this communication process lasts for a predefined duration $\mathcal{T}$, and thus we model it by considering a discrete-time system that evolves over time, \emph{i.e.}, $\mathcal{T} = \{1, 2, \ldots, T\}$. We utilize a three-dimensional (3D) Cartesian coordinate system, and the spatial coordinates of the $i$th UAV and the terrestrial mobile user at time slot $t$ are given by ($x_{i}^U[t], y_{i}^U[t], z_{i}^U[t]$) and ($x^{MU}[t], y^{MU}[t], 0$), respectively. Likewise, the location of the non-associated BS is denoted as ($x^{BS}, y^{BS}, 0$).

\par In the following subsections, we provide detailed descriptions of the models used in this system, including the communication model, UAV movement model, and the random walk model of the terrestrial mobile user, thereby characterizing the key variables of the system. 

\begin{table}[]
\caption{List of Main Notations}
\label{tab:notation}
\begin{tabular}{@{} p{1.2cm} p{7.2cm} @{}}
\toprule
\textbf{Notation}             & \textbf{Definition}                 \\ \midrule
\multicolumn{2}{l}{Notation used in system model} \\
\hline
$N$ & Number of UAVs \\
$\mathcal{N}$ & Set of UAVs \\
$T$ & Number of time slots \\
$\mathcal{T}$ & Set of time slots\\
$d^h_{max}$ & Maximum horizontal flight distance of UAV \\
$d^h_i[t]$ & Horizontal distance of $i$-th UAV flies in time slot $t$ \\
$d^v_{max}$ & Maximum vertical flight distance of UAV \\
$d^v_i[t]$ & Vertical distance of $i$-th UAV flies in time slot $t$ \\
$\psi_i[t]$ & Horizontal direction of $i$-th UAV flies in time slot $t$ \\ 
$K_0$ & Path-loss constant \\
$d_{UM}[t]$ & Distance between UVAA and user in time slot $t$ \\
$d_{BM}[t]$ & Distance between the non-associated BS and user in time slot $t$ \\
$AF[t]$ & The array factor of UVAA in time slot $t$ \\
$g_{UM}[t]$ & Channel power gain between the UAV and user in time slot $t$ \\
$g_{BM}[t]$ & Channel power gain between the non-associated BS and user in time slot $t$ \\
$\Omega_{UM}[t]$ & Small-scale fading between UVAA and user in time slot $t$ \\
$\Omega_{BM}[t]$ & Small-scale fading between the non-associated BS and user in time slot $t$ \\
$K_{UM}$ & Rician factor \\
$G_{UM}[t]$ & Gain of UVAA towards the position of user in time slot $t$ \\
$\Upsilon_{UM}[t]$ & Signal-to-interference-plus-noise ratio for the terrestrial mobile user in time slot $t$ \\
$R_{UM}[t]$ & Achievable rate at user in time slot $t$ \\ 
\hline
\multicolumn{2}{l}{\centering Notation used in reinforcement learning} \\
\hline
$\mathcal{S}$ & State space \\
$s[t]$ & State at time step $t$ \\
$\mathcal{A}$ & Action space \\
$a[t]$ & Action at time step $t$ \\
$\boldsymbol{r}[t]$ & Vectorized reward function at time step $t$\\
$\textbf{A}[t]$ & Vectorized advantage estimator \\
$A^{\omega}[t]$ & Weight-sum advantage estimator \\
$\boldsymbol{\omega}$ & Evenly distributed weight vector \\
$\boldsymbol{V}_{\pi}(s)$ & Multi-objective value function in state $s$ \\
$\gamma$ & Discount factor \\
$\boldsymbol{F}(\pi)$ & The objective value vector of policy $\pi$ \\
$n$ & Number of learning tasks \\
$\mathcal{T}_{task}$ & Set of learning tasks \\
$\Gamma_i$ & The $i$-th learning task in set $\mathcal{T}_{task}$ \\
$EP$ & Set of all non-dominated policies \\
\bottomrule
\end{tabular}
\end{table}

\subsection{Communication Model}
\label{subsec:communication model}

\par In the considered communication model, the signal is first generated by the UVAA, then faded through the channel, and finally decoded by the terrestrial mobile user. This process can be modeled separately as follows.

\subsubsection{UVAA Model}
\label{subsubsec: Array factor of UVAA}

\par The signal distribution of the UVAA is evaluated by using the array factor, which is given by \cite{mozaffari2019}:
\begin{equation}
\begin{split}
    \label{eq.AF}
    A&F[t](\theta, \phi) = \\ &\sum_{i=1}^{N} I_{i}[t] e^{j \left[ k_c(x_{i}^{U}[t] \sin\theta \cos\phi + y_{i}^{U}[t] \sin\theta \sin\phi + z_{i}^{U}[t] \cos\theta) \right]},
\end{split}
\end{equation}

\noindent where $\theta \in [0, \pi]$ and $\phi \in [-\pi, \pi]$ represent the elevation and azimuth angles, respectively, which can be derived from Cartesian coordinates. Moreover, $I_{i}[t]$ indicates the excitation current weight of the $i$th UAV at time slot $t$, and $k_c = 2\pi/\lambda$ is the phase constant, with $\lambda$ denoting the wavelength. Note that the excitation current weights are the complex coefficients that determine the amplitude and phase of the signals transmitted by each UAV in the UVAA, and are also known as beamforming coefficients. As such, the excitation current weight of a UAV can determine its transmit power. From Eq.~(\ref{eq.AF}), the array factor of the UVAA is influenced by both the spatial positions and the excitation current weights of the UAVs. 

\par Following this, the enhancement in gain achieved by the UVAA towards the position of the terrestrial mobile user can be derived through the array factor, which is expressed as follows \cite{sun2022secure}:
\begin{equation}
\label{eq:gain}
    G_{UM}[t] = \frac{4 \pi \left| AF(\theta_{UM}[t], \phi_{UM}[t]) \right|^2 \omega(\theta_{UM}[t], \phi_{UM}[t])^2}{\int_{0}^{2\pi} \int_{0}^{\pi} \left| AF(\theta, \phi) \right|^2 \omega(\theta, \phi)^2 \sin \theta d \theta d \phi} \eta,
\end{equation}

\noindent where $(\theta_{UM}[t], \phi_{UM}[t])$ indicates the direction toward the terrestrial mobile user. Moreover, $\omega(\theta, \phi)$ indicates the magnitude of the far-field beam pattern of each UAV element, and $\eta \in [0, 1]$ represents the antenna array efficiency \cite{Mozaffari2019communicationsandcontrol}.

\subsubsection{Channel Model}

\par To reflect real-world conditions, the wireless channel model between the UVAA and the terrestrial mobile user incorporates both large-scale path loss and small-scale fading. Specifically, the channel power gain between the UVAA and the terrestrial mobile user at time slot $t$ can be expressed as follows:
\begin{equation}
    g_{UM}[t] = K_0 d_{UM}[t]^{-\alpha} \Omega_{UM}[t],
\end{equation}

\noindent where $K_0$ represents the path loss constant, and $d_{UM}[t]$ denotes the distance between the transmitter and the receiver at time slot $t$. Moreover, $\Omega_{UM}[t]$ describes the small-scale fading, modeled as a Rician distribution with $\overline{\Omega}_{UM} = 1$. Consequently, the probability distribution function (PDF) of $\Omega_{UM}$ is expressed as follows~\cite{azari2017}:
\begin{equation}
\begin{split}
    f_{\Omega_{UM}}(\omega) = & \frac{(K_{UM} + 1) e^{-K_{UM}}}{\overline{\Omega}_{UM}} e^{\frac{-(K_{UM} + 1) \omega}{\overline{\Omega}_{UM}}} \\ & \times I_0 \left(2\sqrt{\frac{K_{UM} (K_{UM} + 1) \omega}{\overline{\Omega}_{UM}}} \right); \omega \geq 0,
\end{split}
\end{equation}

\noindent where $K_{UM}$ is the Rician factor, described as the ratio of the power in the LoS component to the power in the non-LoS multipath scatters, and $I_0(\cdot)$ represents the zero-order modified Bessel function of the first kind. Similarly, the channel power gain between the non-associated BS and the terrestrial mobile user at time slot $t$ is given by:
\begin{equation}
    g_{BM}[t] = K_0 d_{BM}[t]^{-\alpha} \Omega_{BM}[t].
\end{equation}

\subsubsection{Transmission Model}

\par When the terrestrial mobile user receives the signal from the UVAA, the unwanted overflow signals from the non-associated BS will cause interference. In this case, the signal-to-interference-plus-noise ratio (SINR) for the terrestrial mobile user at time slot $t$ is given by:
\begin{equation}
\label{eq:SINR}
\Upsilon_{UM}[t] = \frac{P_U[t] G_{UM}[t] g_{UM}[t]}{\sigma^2 + P_B[t] G_{BM} g_{BM}[t]},
\end{equation}

\noindent where $\sigma^2$ represents the noise power, and $P_U[t]$ and $P_B[t]$ represent the total transmit powers of the UVAA and the BS with beamforming function, respectively, $G_{BM}$ is the antenna gain of the sidelobe towards the mobile user, and $g_{BM}[t]$ is the channel gain from the BS to the mobile user. Note that the interference power from a multi-antenna BS employing beamforming is often similar to that from a single-antenna BS~\cite{Dahrouj2010}. This is because the main lobe of the BS is directed towards the intended user~\cite{Javed2023}, and the interference from the BS to the mobile user of UVAA is primarily due to the sidelobe leakage, which is generally low due to the beamforming. Thus, this model can also represent the single antenna BS case. Moreover, each UAV within the UVAA possesses an individual transmit power, and we consider that the maximum transmit power of each UAV is identical. Consequently, the achievable rate for the UVAA-to-terrestrial mobile user link is as follows:
\begin{equation} 
\label{eq:rate}
    R_{UM}[t] = \log_2(1 + \Upsilon_{UM}[t]). 
\end{equation}

\par As illustrated by Eqs.~(\ref{eq:gain}), (\ref{eq:SINR}), and (\ref{eq:rate}), without considering the impact of uncontrollable channel factors, the achievable rate of the UVAA system toward the receiver at each time slot exhibits a positive correlation with the array factor of the UVAA.

\subsection{UAV Movement and Energy Consumption Models}
\label{subsec: UAV Movemnt Model}

\par We consider that the UAVs possess fully controllable mobility to change their 3D positions~\cite{Yao2020Joint3D}. In this case, we denote $\psi_i[t]$ ($0 \leq \psi_i[t] \leq 2\pi$) as the moving direction of the $i$-th UAV in the horizontal plane, and let $d^h_i[t]$ and $d^v_i[t]$ as the horizontal and vertical moving distances of the $i$-th UAV in time slot $t$, respectively. Moreover, the 3D coordinates of the $i$-th UAV in time slot $t$ are given by $(x^U_i[t], y^U_i[t], z^U_i[t])$, and then the 3D coordinates in time slot $t+1$ are given by
\begin{equation}
    \left\{\begin{array}{ll}
        x^U_i[t+1] = x^U_i[t] + d^h_i[t] \cdot \cos \left(\psi_i[t]\right) \\
        y^U_i[t+1] = y^U_i[t] + d^h_i[t] \cdot \sin \left(\psi_i[t]\right) \\
        z^U_i[t+1] = z^U_i[t] + d^v_i[t].
    \end{array}\right.
\end{equation}



\par Based on the aforementioned UAV movement model, we derive the UAV energy consumption model as follows. Specifically, we consider a group of rotary-wing UAVs, and when any UAV $i$ flies at a speed of $v_i$ within a two-dimensional (2D) horizontal plane, its propulsion power consumption is given by~\cite{zeng2019energy}: 
\begin{equation}
\begin{split}
    P(v_i) = & P_B \left( 1 + \frac{3 v_i^2}{v_{tip}^2} \right) \\ &+ P_I \left( \sqrt{1 + \frac{v_i^4}{4v_0^4}} - \frac{v_i^2}{2 v_0^2} \right)^{1/2} + \frac{1}{2} d_0 \rho s A v_i^3, 
\end{split}
\end{equation}
\noindent where $P_B$ and $P_I$ denote the constants corresponding to the blade profile and induced powers during hover, respectively. Furthermore, $v_{tip}, v_0, d_0, \rho, s,$ and $A$ represent the rotor blade's tip speed, the average rotor induced velocity in hover, the fuselage drag ratio, air density, rotor solidity, and rotor disc area, respectively.

\par Following this, by considering UAV climbing and descending actions over time, the 3D energy consumption model of a UAV can be described by using a heuristic closed-form approximation, \textit{i.e.}, \cite{zeng2019accessing}:
\begin{equation}
    \begin{split}
        E_{fly} \approx & \int_0^T P(v_i[t]) dt + \frac{1}{2} m_{UAV} ( v_i[T]^2 - v_i[0]^2) + \\ & m_{UAV} g(H[T] - H[0]), 
    \end{split}
\end{equation}
\noindent where $v_i[t]$ indicates the instantaneous speed of the $i$-th UAV in time slot $t$, $T$ denotes the end time of the flight, $m_{UAV}$ refers to the aircraft mass of a UAV, and $g$ is the gravitational acceleration.

\subsection{Memory-based User Random Mobility Model}
\label{subsec: memory-based user random walk model}

\par To better model the real-world user mobility, we introduce a memory-based random walk model to encapsulate temporal dependencies. Specifically, the current speed and direction of the user are influenced by their previous speed and direction, thereby establishing a correlation between the velocities of the user across successive time slots.

\par We first introduce the Gauss Markov model~\cite{tabassum2019fundamentals} to capture the temporal correlation inherent in the velocity of a terrestrial mobile user. In this model, the velocity of a terrestrial mobile user exhibits temporal correlation and follows a Gauss-Markov stochastic process, \textit{i.e.,}:
\begin{equation}
    v_t = \alpha_g v_{t-1} + ( 1 - \alpha_g ) \mu + \sqrt{1 - \alpha_g^2} \omega_{t-1},
\end{equation}
\noindent where $0 < \alpha_g < 1$ represents the memory level, and $\mu$ denotes the asymptotic mean. Additionally, $\omega_{t}$ signifies an independent, uncorrelated, and stationary Gaussian process with zero mean and variance $\sigma_g^2$, where $\sigma_g$ corresponds to the asymptotic standard deviation. Furthermore, $v_t$ and $v_{t-1}$ denote the velocities in time slots $t$ and $t-1$, respectively. Likewise, let $\Theta$ be the moving direction, which can be derived as follows:
\begin{equation}
    \Theta_t = \alpha_g \Theta_{t-1} + ( 1 - \alpha_g ) \mu + \sqrt{1 - \alpha_g^2} \omega_{t-1}.
\end{equation}

\par Clearly, the parameter $\alpha_g$ modulates the degree of temporal dependency. For instance, for $\alpha_g = 0$, the velocity and direction are determined solely by an independent Gaussian random variable, resulting in entirely stochastic motion. Conversely, if $\alpha_g = 1$, the trajectory becomes linear, where the velocity and direction consistently align with their preceding values. 


%
%
%
\section{Problem Formulation and Analysis}
\label{sec: problem formulation and analysis}

\par In this section, we first formulate the MOP and then present the analyses of the formulated MOP.

\subsection{Problem Formulation}

\par Our primary objective is to maximize the cumulative transmission rate from the UVAA to the terrestrial mobile user by learning the trajectory and potential regularity of the terrestrial mobile user. Meanwhile, the network environment is complicated by time-varying channels and non-associated BSs, which may contribute to significant link instability. To achieve this goal under unstable conditions, we aim to optimize the beam patterns of the UVAA to enhance directivity towards the terrestrial mobile user. Consequently, the UAVs adjust their positions and excitation current weights in real-time. However, the frequent movements of the UAVs increase their energy consumption, potentially reducing the lifespan of the UAV network. Therefore, it is crucial to concurrently address these two conflicting optimization objectives.

\par We define $X = \left[ \mathbb{D}, \mathbb{H}, \mathbb{Z}, \mathbb{I} \right]$ as the decision variables of the optimization problem. Specifically, $\mathbb{D} = \{ \psi_i[t] | \forall i \in \mathcal{N}, \forall t \in \mathcal{T} \}$ refers to the horizontal movement direction of each UAV for all time slots, and $\mathbb{H} = \{ d^h_i[t] | \forall i \in \mathcal{N}, \forall t \in \mathcal{T} \}$ represents the horizontal flight distance of each UAV for all time slots, $\mathbb{Z} = \{ d^v_i[t] | \forall i \in \mathcal{N}, \forall t \in \mathcal{T} \}$ indicates the vertical flight distance of each UAV for all time slots. Moreover, $\mathbb{I} = \{ I_i[t] | \forall i \in \mathcal{N}, \forall t \in \mathcal{T} \}$ refers to the excitation current weights for each UAV for all time slots. Using these variables, we can compute the 3D positions of the UAVs at each time slot. The optimization objectives are subsequently delineated based on these parameters. 


\par \emph{Optimization Objective 1:} The primary optimize objective is to maximize the total achievable rate from the UVAA to the terrestrial mobile user over $T$ time slots, thereby improving the data transmission process. Based on Eqs.~(\ref{eq:SINR}) and~(\ref{eq:rate}), the total achievable rate from the UVAA to the terrestrial mobile user is designed as follows:
\begin{equation}
\label{eq:objective_1}
    f_1 \left( \mathbb{D}, \mathbb{H}, \mathbb{Z}, \mathbb{I} \right) = \sum_{t=1}^T R_{UM}[t].
\end{equation}

\par \emph{Optimization Objective 2:} To enhance the first optimization objective, the UAVs are required to fine-tune their positions frequently. However, the frequent adjustment process will lead to additional motion energy consumption. Thus, to minimize the total motion energy consumption of the UAVs, the second objective function can be expressed as follows:
\begin{equation}
    f_2 \left( \mathbb{D}, \mathbb{H}, \mathbb{Z}\right) = \sum_{t=1}^{T} \sum_{i=1}^{N_{UAV}} E_{i}[t],
\end{equation}

\noindent where $E_i[t]$ represents the energy consumption of the $i$-th UAV in time slot $t$.

\par Note that improving the transmission performance of UVAA increases UAV movement and energy consumption, making it challenging to optimize optimization objectives 1 and 2 simultaneously. Conventional methods like the weighted sum method~\cite{Wang2018} and the constraint method~\cite{pirouz2016computational}, which aggregate or transform the objectives, can simplify the problem but may fail to capture the full trade-offs or limit exploration of the solution space. These methods also struggle in dynamic environments where the importance of each objective can change over time. In contrast, using MOP principles offers several advantages. Specifically, using MOP enables comprehensive exploration of trade-offs, provides flexibility for decision-makers to select solutions based on current priorities, and adapts to the dynamic nature of the system without the need for redefined weights or constraints. Therefore, we aim to use MOP theory to offer a more flexible and comprehensive approach.

\par Accordingly, the MOP based on the aforementioned two optimization objectives is formulated as follows:
\begin{subequations}
\label{eq:formulation}
\begin{align}
    \underset{X}{\max} \ & ( f_1, -f_2 ), \\
    \text{s.t.} \ & 0 \leq I_i[t] \leq 1, \ \forall i \in \mathcal{N}, \forall t \in \mathcal{T}, \label{eq:formulation:subeq1}\\
    & 0 \leq \psi_i[t] \leq 2 \pi, \ \forall i \in \mathcal{N}, \forall t \in \mathcal{T}, \label{eq:formulation:subeq2}\\
    & 0 \leq d^h_{i}[t] \leq d^h_{max}, \ \forall i \in \mathcal{N}, \forall t \in \mathcal{T}, \label{eq:formulation:subeq3}\\
    & -d^v_{max} \leq d_{i}^v[t] \leq d^v_{max}, \ \forall i \in \mathcal{N}, \forall t \in \mathcal{T}, \label{eq:formulation:subeq4}\\
    & L_{min} \leq x_i^U[t] \leq L_{max}, \ \forall i \in \mathcal{N}, \forall t \in \mathcal{T}, \label{eq:formulation:subeq5}\\
    & L_{min} \leq y_i^U[t] \leq L_{max}, \ \forall i \in \mathcal{N}, \forall t \in \mathcal{T}, \label{eq:formulation:subeq6}\\
    & H_{min} \leq z_i^U[t] \leq H_{max}, \ \forall i \in \mathcal{N}, \forall t \in \mathcal{T}, \label{eq:formulation:subeq7}\\
    & d_{(i_1, i_2)}[t] \geq d_{min}, \ \forall i \in \mathcal{N}, \forall t \in \mathcal{T}, \label{eq:formulation:subeq8}
\end{align}
\end{subequations}

\noindent where $X = \{\mathbb{D}, \mathbb{H}, \mathbb{Z}, \mathbb{I}\}$ represents the decision variables of the optimization problem. Moreover, constraint~(\ref{eq:formulation:subeq1}) ensures that the excitation current weight varies between 0 and 1, and constraints~(\ref{eq:formulation:subeq2}),  (\ref{eq:formulation:subeq3}), and (\ref{eq:formulation:subeq4}) confine the horizontal direction, horizontal distance, and vertical distance of the $i$-th UAV, respectively. Furthermore, constraints~(\ref{eq:formulation:subeq5}), (\ref{eq:formulation:subeq6}), and (\ref{eq:formulation:subeq7}) define the movement area of the UAVs. In addition, constraint~(\ref{eq:formulation:subeq8}) ensures that the minimum distance between any two adjacent UAVs at any time slot must exceed $d_{min}$ to prevent collisions.

\subsection{Problem Analysis}

\par In this section, we analyze the formulated MOP as follows.


\par \textit{The formulated MOP is classified as NP-hard since the first optimization objective, as presented in Eq.~(\ref{eq:objective_1}), can be simplified to problems that are known to be NP-hard.} We focus on a single time slot within the first optimization objective and streamline the problem by fixing the positions of the UAVs and the terrestrial mobile user. Consequently, the decision variable can be reduced to solely encompassing the excitation current weight, denoted as $X' = [I_1, I_2, \cdots, I_N ]$. In this case, the simplified first optimization objective denoted as $f_1'$ can be expressed as
    \begin{subequations}
    \begin{align}
        \underset{X'}{\min} \ & f_1' = -R_{UM}[0], \\
        \text{s.t.} \ & g(X') < N, \\
        \ & 0 \leq I_i[t] \leq 1, \ \forall i \in \mathcal{N},
    \end{align}
    \end{subequations}
    \noindent where $g(X') = \sum_{i=1}^N I_i$, in which $N$ is a constant. As such, the expression for $f_1'$ represents a nonlinear knapsack problem, which is proven to be NP-hard. As a result, the original optimization problem, as depicted in Eq.~(\ref{eq:formulation}), is also NP-hard, given that it is more complex than  $f_1'$.

\par \textit{Trade-offs exist between the optimization objectives considered in the formulated MOP.} Specifically, to enhance the directivity and gain of the beam pattern (\emph{i.e.}, to maximize $f_1$), the UAVs must navigate to the optimal position according to the location of the user during each time slot. However, this frequent repositioning entails significant energy consumption for the UAVs, leading to an increase in $f_2$. Clearly, it becomes evident that the two optimization objectives of the formulated MOP exhibit a conflicting relationship.

\par \textit{The formulated MOP is both dynamic and regular.} Specifically, the considered environment is changing rapidly, including time-varying channels and interference from non-associated BS, resulting in the instability of communication links. Furthermore, given the random mobility patterns of users, the UAVs need to constantly fly to the optimal position based on the trajectory of the user. Consequently, this aspect renders the formulated MOP inherently dynamic. Additionally, in realistic scenarios, the movement patterns of terrestrial mobile users are often goal-oriented, suggesting that the future speed and direction of the user may be related to their current speed and direction. This observation implies a degree of regularity in the formulated MOP.

\par The formulated MOP requires real-time decision-making and has long-term optimization objectives. Specifically, the mobility of the ground user and time-varying wireless channels introduce significant uncertainty. Thus, the UAVs need to adapt their trajectories in real-time to maintain optimal communication links. Moreover, in the considered scenario, the actions of UAVs made at each time step affect future achievable transmission rates and energy efficiency of UAVs, necessitating a method that considers long-term optimization objectives.


\par Given its hardness and complexity, the formulated MOP poses significant challenges for resolution using common optimization algorithms. In the following sections, we propose a novel evolutionary DRL algorithm to address the formulated problem.

\section{Algorithm}
\label{sec:algorithm}

\par In this section, we initially introduce the preliminaries of DRL. We then transform the formulated MOP into a multi-objective Markov decision process (MOMDP). Finally, we detail the proposed EMOPPO-VLH.

%
%
\subsection{The Preliminaries of DRL}

\par We first present the motivations for using DRL and then introduce the preliminaries of MOMDP. 

\subsubsection{Motivations for Using DRL}

\par Due to the properties of the formulated MOP, the various conventional optimization methods are not suitable for solving it. \textbf{First}, the NP-hard nature of the problem and the non-linear relationships between objectives and decision variables make it difficult to find an optimal solution by using conventional methods (\textit{e.g.}, exhaustive approach or convex optimization~\cite{Nievergelt2000, boyd2004convex}). \textbf{Second}, the formulated optimization problem is a long-term sequence decision-making problem, and the considered scenario may also confront uncertainty and dynamic conditions. This requires balancing short-term gains with long-term objectives, causing evolutionary algorithms with low performance~\cite{Bliss2014}. \textbf{Finally},  the massive solution space and complex energy consumption model of UAVs make it infeasible to design a high-performance and cooperative online algorithm~\cite{nikolos2003evolutionary} (\textit{e.g.}, the algorithm with a tight competitive ratio). 

\par In this case, DRL offers significant advantages for such optimizations, especially in dynamic environments. Specifically, DRL is a machine learning paradigm that combines reinforcement learning with deep neural networks to solve complex decision-making problems in dynamic environments~\cite{Nguyen2020}. At the core of DRL is the Markov decision process (MDP), which provides a mathematical framework for modeling sequential decision-making under uncertainty~\cite{guo2022real}. In an MDP, an agent interacts with an environment in discrete time steps, making decisions that maximize cumulative rewards. In this case, the DRL agent seeks to learn an optimal policy $\pi^*$ that maximizes the expected cumulative reward over time by interacting with the environment and updating its policy based on the rewards received.

\par As such, DRL demonstrates significant potential for addressing the formulated MOP. This is particularly relevant for our problem of optimizing UAV trajectories and excitation current weights, which requires handling dynamic environmental changes, making real-time decisions, and achieving long-term optimization objectives in an NP-hard solution space. In particular, DRL continuously learns from its environment through trial and error to adapt to dynamic and unpredictable conditions~\cite{ZhangSurvey}. This enables the refinement of strategies in real-time, thereby making DRL particularly effective in environments where conditions and objectives are continuously evolving. Furthermore, the ability of DRL to optimize for long-term rewards allows it to balance competing objectives by considering future outcomes, rather than focusing solely on short-term gains. Therefore, the strong generalization capabilities of DRL and its ability to learn under uncertainty make it particularly well-suited for complex and real-time decision-making that requires continuous adaptation.

\par Accordingly, we aim to use multi-objective optimization theory to model the formulated MOP and use DRL to solve it. Unlike single-objective optimization problems, it is challenging to find an optimal policy that simultaneously maximizes all objectives. Therefore, MOPs aim to identify a non-dominated set of solutions, namely, the Pareto set. In the following, we introduce the MOMDP for enabling multi-objective DRL.

%
%
\subsubsection{MOMDP}

\par The formulated MOP is a multi-objective sequential decision problem that can be formulated as an MOMDP \cite{xu2020prediction}. Specifically, the MOMDP extends the MDP and is denoted by the tuple $\langle \mathcal{S}, \mathcal{A}, \mathcal{P}, \mathbf{R}, \gamma, \mathcal{D} \rangle$. Within this tuple, $\mathcal{S}$, $\mathcal{A}$, $\mathcal{P} (s' | s, a)$, and $\mathbf{R} = (r_1, \ldots, r_M)$ denote the state space, action space, state transition probability, and vector of reward functions, respectively, where $r_m$ is the reward for each of the considered $M$ objectives. Moreover, $\gamma \in [0, 1)$ represents the discount factor, and $\mathcal{D}$ represents the initial state distribution. The details of the agent interaction with MOMDP can be found in Appendix A.1.

\par Based on MOMDP models, we will transform the formulated MOP into an MOMDP and solve it by using a DRL-based method in the following section.

\subsection{MOMDP Formulation}

\par To solve the formulated MOP using DRL-based algorithms, the key elements of the considered MOMDP can be described as follows:

\subsubsection{State Space}

\par State space contains the environment information of the formulated problem. Given that both the UAV and the terrestrial mobile user are equipped with positioning devices (\textit{e.g.}, global positioning system (GPS)), the locations of the UAVs and the terrestrial mobile user can be easily accessed. The state consists of the positions of all UAVs as well as the location of the user, which is defined as follows:
\begin{equation}
\begin{aligned}
    \mathcal{S} =  \{s[t] | s[t] = \left ( \{\mathbf{c}^U_i[t]\}_{i \in \mathcal{N}}, \mathbf{c}^M[t]\right), \forall t \in \mathcal{T}  \},
\end{aligned}
\end{equation}
    
\noindent where $\mathbf{c}^U_i = (x^U_i[t], y^U_i[t], z^U_i[t])$ and $\mathbf{c}^M[t] = (x^M[t], y^M[t], 0)$ are the coordinates of the $i$-th UAV and the terrestrial mobile user at time slot $t$, respectively.
    
\subsubsection{Action Space}

\par The action space can represent the decision variables of the formulated problem. According to the observed state, the central controller chooses the horizontal direction $\psi_i[t]$, the horizontal distance $d^h_i[t]$, the vertical distance $d^v_i[t]$, and the excitation current weight $I_i[t]$ for each UAV at time $t$ to perform UVAA. Hence, the action can be defined by
\begin{equation}
\begin{aligned}
    \mathcal{A} = \{ a[t] | a[t] = \big( \{I_i[t]\}_{i \in \mathcal{N}}, \{\psi_i[t]\}_{i \in \mathcal{N}}, \{d^h_i[t]\}_{i \in \mathcal{N}}, \\ \{d^v_i[t]\}_{i \in \mathcal{N}} \big), \forall t \in \mathcal{T} \}.
\end{aligned}
\end{equation}

\noindent Note that we define UAV actions as directions instead of 3D Cartesian coordinates since this manner can effectively capture UAV temporal dynamics, align with practical control schemes, and simplify the action space, which has the potential to facilitate more efficient learning performance of DRL. This manner is also adopted by several existing works~\cite{Zhang2021, Mei2022, Bayerlein2021, Li2020a}.

\subsubsection{Reward Function}

\par In a DRL-based framework, the environment provides immediate feedback in the form of rewards subsequent to the execution of an action. The agent depends on the reward to modify its actions and develop optimal policies for maximizing the reward. Therefore, the design of the reward system is crucial in enhancing the performance of the system. In this study, we aim to maximize the total achievable rate while minimizing the total energy consumption of the UAVs. Unlike the scalar reward in a single-objective MDP, the reward structure in an MOMDP is a vector. As such, we define the reward function as follows:
\begin{equation}
    \boldsymbol{r}[t] = ( r^R[t], r^E[t]) = 
    \begin{cases}
        ( R_{UM}[t], -\varepsilon_1 E[t]),& \text{if $\mathbbm{1}[t] = 1$} \\
        (\varepsilon_2 R_{UM}[t], -\varepsilon_1 \varepsilon_3 E[t]),& \text{otherwise} 
    \end{cases}
\end{equation}

\noindent where $r^R[t]$ and $r^E[t]$ are the scaling rewards corresponding to $R_{UM}[t]$ and $E[t]$ in time slot $t$, respectively. The indicator variable $\mathbbm{1}[t]$ is assigned a value of 0 if the UAVs attempt to fly outside the designated area or if a collision occurs between adjacent UAVs at time $t$, and $\mathbbm{1}$ is equal to 1, otherwise. Moreover, coefficients $\varepsilon_1$, $\varepsilon_2$, and $\varepsilon_3$ are implemented to penalize the reward in cases where the UAVs breach the area restriction (as outlined in constraints (\ref{eq:formulation:subeq5})-(\ref{eq:formulation:subeq7})) or the collision avoidance restriction (specified in constraint (\ref{eq:formulation:subeq8})).

\subsection{Multi-Objective Proximal Policy Optimization Task}

\par Based on the aforementioned MOMDP, we define a learning task for solving our problem as a tuple $\Gamma = \langle \boldsymbol{\omega}, \pi_{\theta} \rangle$, where $\boldsymbol{\omega} \left ( \sum_m^M \omega_m = 1 \right)$ denotes a weight vector, and $\pi_{\theta}$ denotes a policy to be optimized. The goal of the considered multi-objective DRL method is to maximize the weighted-sum reward, \emph{i.e.}, 
\begin{equation}
    J(\theta, \boldsymbol{\omega}) = \sum_{m=1}^M \omega_m f_m(\pi) = \sum_{m=1}^M \omega_m J_m^{\pi}.
\end{equation} 

\par Note that in the learning process of our proposed EMOPPO-VLH, a policy will be optimized by using different weights. To this end, proximal policy optimization (PPO)~\cite{schulman2017proximal}, which is a powerful DRL approach, can be employed to handle the multi-objective weight-sum task with high-dimension action space. Specifically, PPO is an actor-critic, on-policy, and policy-gradient algorithm extended from trust region policy optimization (TRPO). PPO aims to prevent significant deviations of new policies from old ones by adopting a clipped surrogate objective, which is given by
\begin{equation}
\label{eq:clip}
\begin{aligned}
    & J^{clip}(\theta, \boldsymbol{\omega}) = \\& \mathbb{E} \left[ \sum_{t=1}^T \min \left( r[t](\theta) A^{\omega}[t], \text{clip}(r[t](\theta), 1 - \epsilon, 1 + \epsilon) A^{\omega}[t] \right) \right],
\end{aligned}    
\end{equation}


\noindent where $\epsilon$ denotes a hyperparameter used to govern the clip range, and $r[t](\theta) = {\pi_{\theta} (a[t] | s[t])} / {\pi_{\theta_{old}} (a[t] | s[t])}$ denotes the probability ratio. Moreover, $A^{\omega}[t] = \boldsymbol{\omega A}[t]$ is the weight-sum advantage estimator, where $\boldsymbol{A}[t]$ is the vectorized advantage estimator, defined as follows:
\begin{equation}
    \label{eq:advantage}
    \boldsymbol{A}[t] = \sum_{k=0}^{T-t+1} (\gamma \lambda)^k (\boldsymbol{r}[t+k] + \gamma \boldsymbol{V}_{\pi}(s[t+k+1]) - \boldsymbol{V}_{\pi}(s[t+k])),
\end{equation}

\noindent where $\lambda \in [0, 1]$ controls the balance between bias and variance. Note that the value function from the single-objective PPO algorithm can not enable it to handle multiple optimization objectives simultaneously. As such, $\boldsymbol{V}_{\pi}(s)$ represents the proposed \textbf{vectorized value function}, which associates the state $s$ with a vector of expected returns given the policy $\pi$. By vectorizing the value function, the value function from a previous training process can be directly adapted to optimize the same policy with updated weights. As such, the value loss function used for updating the parameters of the value network is given by
\begin{equation}
    \label{eq:loss}
    J^V(\theta) = \mathbb{E} \left[ \sum_{t=1}^T \| \boldsymbol{V}_{\pi}(s) - \hat{\boldsymbol{V}}_{\pi}(s) \|^2 \right],
\end{equation}

\noindent where $\hat{\boldsymbol{V}}(s) = \boldsymbol{r}[t] + \gamma \boldsymbol{V}_{\pi}(s[t+1])$ is the target value function. 

\par Despite the effective performance of the PPO algorithm in UAV-assisted networks, it faces several challenges within the considered environment. First, time-varying channel fading leads to uncertain network state transitions, increasing the learning uncertainties and reducing the accuracy. Additionally, the current speed and direction of the user are frequently influenced by their historical behavior, such as moving toward a predetermined destination. This pattern of time-dependent movement is not effectively captured by purely fully-connected neural networks. To address the aforementioned issues, we propose the integration of an LSTM architecture for exploiting the temporal sequence of user movements.

\subsection{The Proposed EMOPPO-VLH Method}
\label{subsec:IEMORL}

\begin{figure*}[t]
    \centering
    \centerline{\includegraphics[scale=0.65]{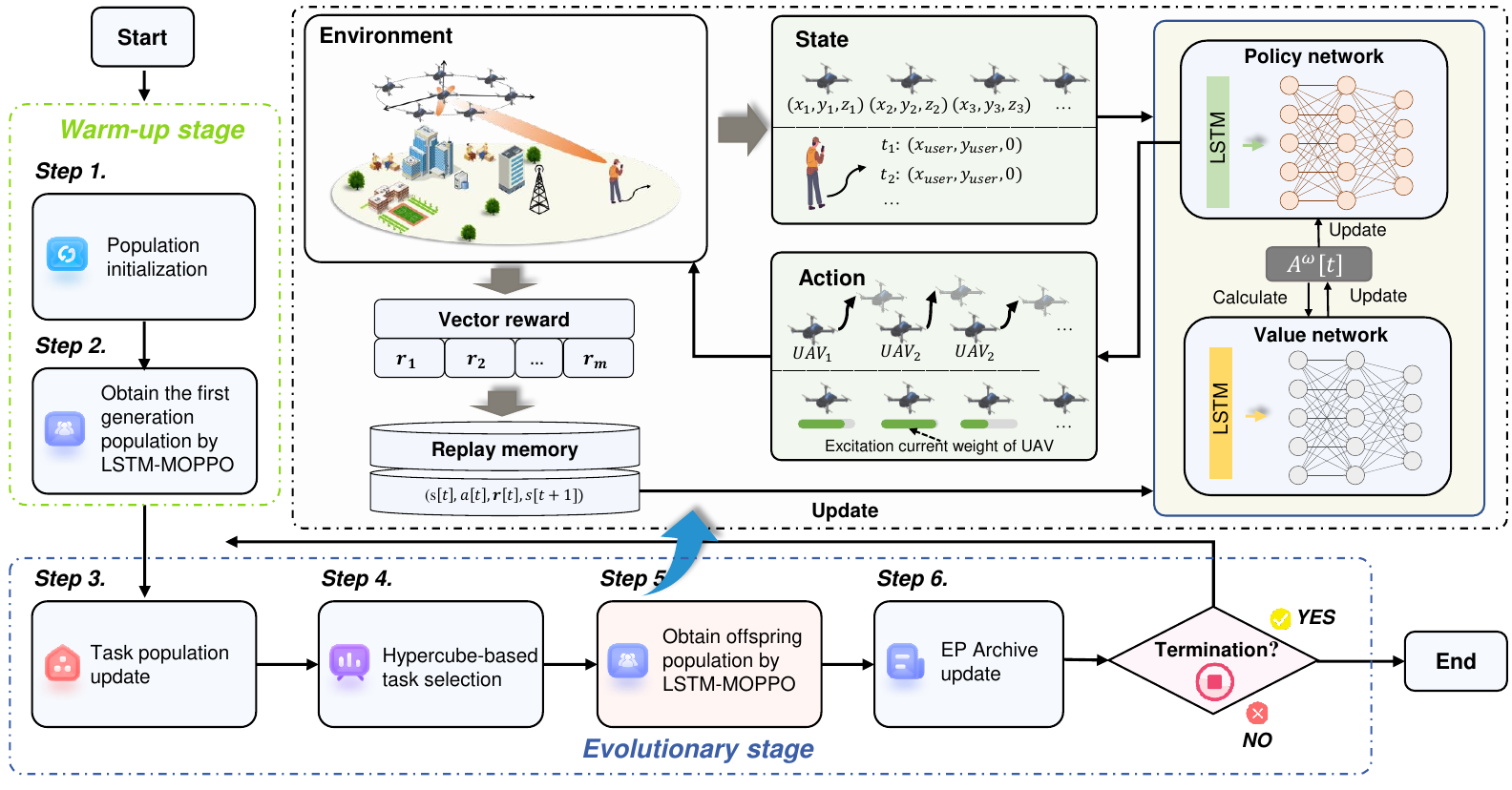}}
    \caption{The algorithmic framework of EMOPPO-VLH is initiated with a warm-up stage, designed to generate a high-quality primary population. Subsequently, EMOPPO-VLH advances to the evolutionary stage, which encompasses task population update, task selection, acquisition of offspring population, and EP archive update. The tasks selected during this stage are optimized by using the LSTM-MOPPO algorithm, resulting in a new generation of offspring. The architecture of LSTM-MOPPO is represented by the part of a black dashed line in the diagram.}
    \label{fig:MODRL}
\end{figure*}

\par We propose an EMOPPO-VLH framework aimed at obtaining a set of Pareto optimal policies. As shown in Fig.~\ref{fig:MODRL}, this framework incorporates multiple learning tasks, each guided by a DRL policy. These learning tasks evolve over iterations through multi-objective optimization and are improved via DRL training. Specifically, our proposed method begins with a warm-up stage. In this stage, we obtain $n$ randomly initialized policies and $n$ weight vectors, which are evenly distributed. These constitute $n$ sets of learning tasks. The first generation of policy populations is generated by executing LSTM-MOPPO to optimize each learning task. Following this, EMOPPO-VLH proceeds to the evolutionary stage. In each generation of the evolutionary stage, a hyper-sphere-based task selection method is proposed and Pareto dominance~\cite {li2021physical} is introduced to select $n$ learning tasks aimed at improving the diversity and quality of the Pareto set. Next, the selected learning tasks are optimized by using LSTM-MOPPO to generate new offspring policies. Then, the external Pareto archive and the policy population are updated by using the offspring population. The evolutionary stage continues iteratively until the predefined number of generations is achieved. Note that the predefined number of generations is determined based on whether the EMOPPO-VLH converges stably.

\par The pseudo-code of EMOPPO-VLH is provided in Algorithm~\ref{alg:IEMORL} and the detailed descriptions of the LSTM network structure and the evolutionary learning process are presented below.

\subsubsection{LSTM Network Structure}

\par As mentioned above, the time-varying channel fading and user movement can introduce significant dynamics into the considered system and MOMDP. These uncertainties can lead to the reduced learning accuracy and increased difficulty. Moreover, the current direction and velocity of the user often depend on their previous direction and velocity. In conventional PPO, the actor and critic networks use fully connected neural networks, which operate under the assumption that all inputs are independent of one another. Consequently, it is challenging for traditional neural network structures to capture the temporal dependency of sequential observations over time.

\par To overcome this issue, we seek to use LSTM networks as the structure for both the actor and critic networks, which enable the capture of hidden user movement patterns. Specifically, LSTM is a variant of recurrent neural networks but can more efficiently capture temporal dependencies through its specialized gate mechanisms. As such, we replace the first fully connected layer with an LSTM layer. The LSTM cell comprises an input gate, an output gate, and a forget gate, and their structures can be detailed as follows~\cite{Chen2024information}:

\begin{itemize}
    \item \textbf{Forget gate:} The forget gate determines the amount of previous information to discard by using a sigmoid function. 

    \item \textbf{Input gate:} The input gate decides which information to retain in the cell state.

    \item \textbf{Output gate:} The output gate determines the information to be outputted. Specifically, the new cell state is obtained by combining the current cell state with the output from the input gate, subsequently generating the LSTM output. Moreover, the output gate also consists of a sigmoid layer and a tanh layer. 

\end{itemize}

\par As such, the neural network of the proposed MOPPO can excel at retaining long-term relevant states and discarding irrelevant ones. This capability gives MOPPO a significant advantage in extracting temporal features from the environment, particularly in scenarios where user movement patterns and channel variations exhibit strong temporal correlations.

\begin{algorithm}[t]
    \caption{EMOPPO-VLH}
    \label{alg:IEMORL}
    \KwIn{$n$ Learning tasks, warm-up iterations $n_{warm}$, task iterations $n_{evo}$, evolution generations $G$}
    \tcc{\textbf{Warm-up stage}}
    Initialize population $P = \emptyset$ and external Pareto archive $EP = \emptyset$; \\
    
    Generate $n$ evenly distributed weight vectors $\mathcal{W} = \{\boldsymbol{\omega}_1, \dots, \boldsymbol{\omega}_n\}$; \\
    
    Initialize $n$ policy networks $\{\pi_{\theta_1}, \dots, \pi_{\theta_n}\}$ with LSTM architecture; \\

    Initialize $n$ value networks $\{ \boldsymbol{V}_{\pi_{\theta_1}}, \dots, \boldsymbol{V}_{\pi_{\theta_n}} \}$ with LSTM architecture; \\

    Generate task set $\mathcal{T}_{task} = \{ \Gamma_1, \dots, \Gamma_n \}$, where $\Gamma_i = \langle \boldsymbol{\omega_i}, \pi_{\theta_i}, \boldsymbol{V}_{\pi_{\theta_i}} \rangle$\; 

    $P' \leftarrow$ MMPPO($\mathcal{T}_{task}, n_{warm}$)\tcp*{Obtaining the initial population} 

    Update $EP$ by using Pareto dominance\;
    
    \tcc{\textbf{Evolutionary stage}}

    \For {\textit{generation} = $1$ to $G$} { 

        $P \leftarrow$ TPU($P$, $P'$)\tcp*{Updating the task population utilizing Algorithm 3}
    
        $\mathcal{T}_{task} \leftarrow $ TaskSelection($\mathcal{W}$, $P$)\tcp*{Selecting the learning tasks utilizing Algorithm 4}

        $P' \leftarrow$ MMPPO($\mathcal{T}_{task}, n_{evo}$)\tcp*{Obtaining the offspring population utilizing Algorithm 2}

        Update $EP$ by using Pareto dominance\; 

    }
    \KwOut{Pareto archive $EP$}
\end{algorithm}

\subsubsection{Evolutionary Deep Reinforcement Learning Process}

\par The proposed EMOPPO-VLH consists of two main stages, which are the warm-up and evolutionary stages, and their details are provided as follows.

\par \textbf{\textit{Warm-up Stage}:} The algorithm initiates with a warm-up stage. Within this stage, a set of $n$ policies is stochastically generated, each being assigned a corresponding weight. Although these policies share identical state spaces, action spaces, and reward functions, their distinct weight vectors and neural network parameters result in markedly different offspring when the LSTM-MOPPO algorithm is executed. The procedure for task generation in the warm-up stage can be described as follows.

\par Initially, we generate $n$ non-negative weight vectors $\{\boldsymbol{\omega}_1, \dots, \boldsymbol{\omega}_n\}$ that are evenly distributed, with the constraint that $\sum_j \omega_{i, j} = 1$ for $1 \leq i \leq n$. Note that these weight vectors are used to combine the vector reward function into a single scalar reward for training the tasks via LSTM-MOPPO. 
Subsequently, we randomly initialize $n$ policy networks, $\{ \pi_{\theta_1}, \dots, \pi_{\theta_n} \}$, and $n$ multi-objective value networks, $\{ \boldsymbol{V}_{\pi_{\theta_1}}, \dots, \boldsymbol{V}_{\pi_{\theta_n}} \}$, each utilizing an LSTM-based architecture. We then represent the set of learning tasks as $\mathcal{T}_{task} = \{ \Gamma_1, \dots, \Gamma_n \}$, where each task $\Gamma_i$ is defined by the triplet $\langle \boldsymbol{\omega}_i, \pi_{\theta_i}, \boldsymbol{V}_{\pi_{\theta_i}} \rangle$.

\par Finally, each task from $\mathcal{T}_{task}$ undergoes optimization through the LSTM-based multi-objective PPO, as detailed in Algorithm~\ref{alg:lstm-ppo}, for a predetermined number of iterations. The derived policies then constitute the initial generation of the policy population.

\begin{algorithm}[t]
    \caption{LSTM-based multi-objective PPO (LSTM-MOPPO)}
    \label{alg:lstm-ppo}
    \KwIn{Task set $\mathcal{T}_{task}$, number of iterations $n_{iter}$}

    Initialized offspring population $P' = \emptyset$;
    
    \tcp{Each learning task is optimized by using LSTM-MOPPO.}
    \For {$\Gamma = \langle \boldsymbol{\omega}_i, \pi_{\theta_i}, \boldsymbol{V}_{\pi_{\theta_i}} \rangle \in \mathcal{T}_{task}$} {
        \For{$j = 1$ to $n_{iter}$} {
            Collect trajectories by executing the LSTM-augmented policy $\pi_{\theta_i}$;
            
            Calculate the vectorized advantage estimator $\boldsymbol{A}[t]$ by Eq.~(\ref{eq:advantage});

            Calculate the weight-sum advantage estimator $A^{\omega_i}[t] = \boldsymbol{\omega_i A}[t]$;

            Optimize policy network's parameter $\theta_i$ by Eq.~(\ref{eq:clip}) for several epochs;

            Optimize the value network $\boldsymbol{V_{\pi_{\theta_i}}}$ by Eq.~(\ref{eq:loss});

            Collect the updated new task $\Gamma_i$ in $P'$;
        }
    }
    \KwOut{Offspring population $P'$}
\end{algorithm}

\par The design of an effective method for generating a high-quality offspring population during the evolutionary process is paramount. However, the original MOPPO algorithm falls short in capturing the temporal dependencies across various time slots in dynamic environments. To address this, as aforementioned, we integrate an LSTM network to improve the exploration capabilities of the algorithm. Moreover, the original MOPPO retains only tasks solely after $n_{iter}$ iterations in the offspring population, potentially overlooking latent beneficial tasks. By preserving all new tasks after each iteration, we augment the diversity of the offspring population. Explicitly, running our LSTM-MOPPO can generate $n \cdot n_{iter}$ offspring, offering both better exploratory capacity and larger diversity. As a result, the LSTM-MOPPO consistently produces a more qualitative offspring population~\cite{song2022evolutionary}, \cite{Ferdowsi2021neural}.

\par The warm-up stage can generate a set of policies residing in the high-performance region, thereby reducing the noise and uncertainty in the learning process.


\par \textbf{\textit{Evolutionary Stage}:} After completing the warm-up stage and obtaining the initial population, the evolution stage of the algorithm is initiated. Specifically, the policy population $P$ is updated by using the resultant offspring population $P'$, as shown in Algorithm~\ref{alg:TPU}. For the population update, we employ the performance buffer strategy presented in \cite{schulz2018interactive}, which ensures both the preservation of performance and the promotion of diversity within the population. Let $B_{num}$ and $B_{size}$ represent the total number of buffers and the capacity of each buffer, respectively. The performance space is divided into $B_{num}$ buffers, with each buffer capable of storing up to $B_{size}$ learning tasks. The position of policy $\pi_{\theta}$ in the performance space is determined based on the objective vector $\boldsymbol{F}(\pi_{\theta})$ and the reference point $\boldsymbol{Z}_{ref}$. The task associated with $\pi_{\theta}$ is stored in the buffer that lies closest in proximity.

\par
\begin{figure}[t]
	\centerline{\includegraphics[width=3.5in]{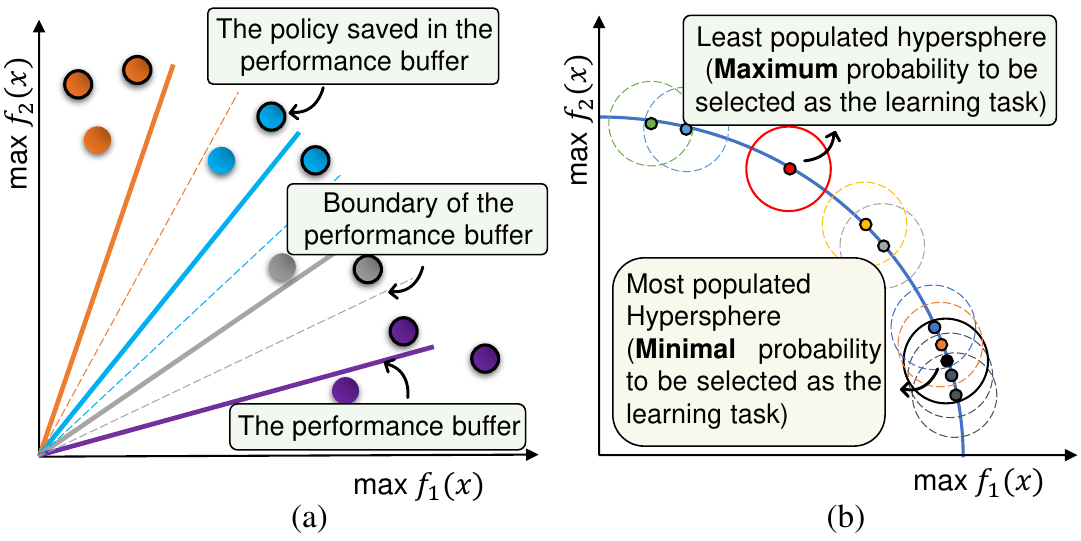}}
	\caption{An illustrative example of performance buffer and hyper-sphere-based task selection strategies. (a) Performance buffer strategies: The lines emanating from the origin represent the buffer. Each circle denotes an objective value calculated by the corresponding policy. Circles outlined in black represent policies that are preserved in the performance buffer. (b) Hyper-sphere-based task selection strategy: The circle around the dot symbolizes the sub-hyper-sphere and the fewer policies it contains, the higher the probability that a corresponding strategy will be selected as the learning task.}
	\label{fig:buffer}
\end{figure}

 If the number of tasks in a buffer exceeds $B_{size}$, we retain only $B_{size}$ learning tasks that have the maximum distance from the reference point $\boldsymbol{Z}_{ref}$, as shown in Fig.~\ref{fig:buffer}(a). Consequently, the updated task population encompasses all learning tasks present in the performance buffer.

\begin{algorithm}[t]
    \caption{Task Population Update (TPU)}
    \label{alg:TPU}
    \KwIn{Task population $P$, offspring population $P'$, reference point $\boldsymbol{Z}_{ref}$, the number of buffer $B_{num}$, the size of buffer $B_{size}$}

    Initialize performance buffer $\mathcal{B}_i = \emptyset, i = 1, \dots B_{num}$\;

    Generate $B_{num}$ evenly distributed weight vectors $\{ \boldsymbol{\omega}_1, \dots, \boldsymbol{\omega}_{B_{num}} \}$\;

    \For(\tcp*[f]{Store the tasks in the performance buffer}){$\Gamma = \langle \boldsymbol{\omega}, \pi_{\theta}, \boldsymbol{V}_{\pi_{\theta}} \rangle \in \{ P \cup P' \}$} {

        Calculate objective vector $\boldsymbol{F}(\pi_{\theta})$\;

        Set $\boldsymbol{F}_{ref} = \boldsymbol{F}(\pi_{\theta}) - \boldsymbol{Z}_{ref}$\;

        Determine the buffer index $\hat{j} = \arg \max_{j = 1, \dots, B_{num}} \{ \frac{\boldsymbol{\omega}_j \cdot \boldsymbol{F}_{ref}}{||\boldsymbol{\omega}_j||} \}$\;

        Store task $\Gamma$ in $\mathcal{B}_{\hat{j}}$\;

        \If(\tcp*[f]{Retain only the first $B_{size}$ tasks}){$|\mathcal{B}_{\hat{j}}| > B_{size}$} {
            Calculate distance between $\boldsymbol{F}(\pi_{\theta})$ and $\boldsymbol{Z}_{ref}$\;

            Retain only the $B_{size}$ tasks exhibiting the greatest distances\;
        }
        
    }

    Update population $P_{new} = \bigcup_{i=1}^{B_{num}} \mathcal{B}_i$\;
    
    \KwOut{updated population $P_{new}$}
\end{algorithm}


\par In each generation, we employ a hyper-sphere-based task selection strategy as depicted in Algorithm~\ref{alg:taskSelection}, aiming to select $n$ learning tasks from $P$ to enhance the diversity of the Pareto set. Initially, the objective vector $\boldsymbol{F}(\pi_{\theta_j})$ for every task $\Gamma_j \in P$, where $j = \{1, \dots, |P|$\}, is computed. For each weight vector $\boldsymbol{\omega}_i \in \mathcal{W}$, we compute the weighted value of each objective vector by using $\boldsymbol{\omega}_i$ as the weighting factor. The tasks corresponding to the highest $k_{can}$ values are extracted to constitute the candidate task set $\mathcal{T}_{can}$. Subsequently, we define a hyper-sphere to encompass all policies in $\mathcal{T}_{can}$ and partition the hyper-sphere into equivalent sub-hyper-spheres. After counting the number $N_i$ of policies within each sub-hyper-sphere, we compute the selection probability of a task using $\mathcal{P} = c / N_i$, where $c$ is a constant greater than one. Finally, the $n$ selected tasks are incorporated into $\mathcal{T}_{task}$, and we generate the offspring $P'$ by executing the LSTM-MOPPO algorithm, using $\mathcal{T}_{task}$ and $n_{evo}$ as input. Here $n_{evo}$ denotes the predetermined number of iterations designated for the evolutionary stage. This selection mechanism is visualized in Fig.~\ref{fig:buffer}(b).

\par Note that parameter $k_{can}$ represents a crucial balance between exploration and exploitation within the algorithm's framework. Specifically, when $k_{can}=1$, each weight vector selects only the most optimal learning task from $\mathcal{P}$. In contrast, for $k_{can} = |\mathcal{P}|$, the selection process for each weight vector adheres strictly to the hyper-sphere configuration, where the learning task associated with the sparsest hyper-sphere is accorded the highest probability of selection.


\begin{algorithm} [t]
    \caption{Hyper-sphere-based Task Selection}
    \label{alg:taskSelection}
    \KwIn{Weight vectors $\mathcal{W}$, task population $P$, the number of candidate task $k$}

    Calculate objective vector $\boldsymbol{F} \left( \pi_j \right)$ of policy $\pi_{\theta_j}$ of each task $\Gamma_j \in P$\;

    \For{$\boldsymbol{\omega}_i \in \mathcal{W}$} {

        Initialize an empty set $temp = \emptyset$\;
        
        \For{$j = 1$ to $|P|$} {
        
            Append $value = \boldsymbol{\omega}_i \cdot \boldsymbol{F} \left( \pi_j \right)$ to $temp$\;

        }

        Sort $temp$ in descending order, and extract the tasks corresponding to the top $k$ values to form the candidate task set $\mathcal{T}_{can}$\;

        Calculate the number $N_i$ of policies present in each hyper-sphere within $\mathcal{T}_{can}$\;

        Select task $\hat{\Gamma}$ by a roulette-wheel mechanism with the probability $\mathcal{P} = c / N_i$ for each task in $\mathcal{T}_{can}$\;

        Update the weight vector of task $\hat{\Gamma}$ to be $\boldsymbol{\omega}_i$\;
        
        Add task $\Gamma$ to $\mathcal{T}_{task}$\;
    }
    
    \KwOut{Selected task set $\mathcal{T}_{task}$}
\end{algorithm}

\par The evolutionary stage terminates upon reaching a specific number of iterations. Throughout this stage, an external Pareto archive is utilized to preserve all non-dominated policies, which will function as an approximation of the Pareto optimal policies for the formulated MOP.

\subsection{Practical Implementation of EMOPPO-VLH-based UAV Management}

\par In this subsection, we provide a detailed procedure for applying our method in practical scenarios, particularly regarding the distinction between the training phase and the deployment phase of our algorithm

\subsubsection{Training Phase}

\par During the training phase, the computation of reward values and the evolutionary learning process require information from all agents, which demands significant computational resources. Therefore, we recommend a centralized implementation of our algorithm, where training occurs on a central controller with sufficient computing power. In this case, akin to the previous works~\cite{10629203, 10574401, 10275021, Guo2024}, the proposed method interacts with a simulation environment constructed by using mathematical models, rather than relying on pre-collected real-world data. The proposed method collects key information from the simulation environment, such as data rates of mobile users, UAV energy consumption, and the locations of users and UAVs, to guide the learning process and optimize policies. 

\par Note that this simulation environment is built by using state-of-the-art models, such as the memory-based user random mobility model, the Rician channel model, and the UAV energy consumption model, all derived from real-world data to ensure realistic simulations. As a result, the policy trained in this environment can be effectively applied to real-world scenarios. Once the algorithm converges, the trained DRL policy can then be deployed in practical environments. Note that, as specified in Algorithm~\ref{alg:IEMORL}, the proposed algorithm produces a set of candidate policies (\emph{i.e.} Pareto set). All the candidate policies are valuable and represent different trade-offs between the two objectives. 

\subsubsection{Deployment Phase}

\par During the deployment phase, the well-trained DRL policy receives the state information from the real environment to make the decisions. Note that these DRL policies can be executed by the central controller within the designed system. The central controller can readily access location information from UAVs and terrestrial mobile users, as they are equipped with positioning devices. During this phase, the algorithm no longer requires real-time data on data rates and UAV energy consumption to calculate rewards. Instead, the algorithm utilizes the trained policy to make control decisions aligned with the current preference based on the current state information, such as the locations of the user and UAVs. Note that periodic intermittent data sampling may be conducted to fine-tune the simulation environment, avoiding the necessity of continuous real-time data exchange. As such, we can disregard the communication overhead between the central controller and the UAVs, because the data size related to the locations of the UAVs is small. 

\subsubsection{Environment Changes}

\par In the case of environment changes, the previously trained neural networks can adjust their parameters based on the training results in the new environment. Specifically, the administrator can modify key parameters of the simulation environment (\textit{e.g.}, UAV masses, user patterns) and retrain the DRL algorithm. Efficient techniques such as transfer learning~\cite{Zhuang2021} can help ensure quick convergence. The introduced LSTM, hyper-sphere-based task selection method, and value function extension also boost convergence speed. Based on this, leveraging edge computing and incremental learning allows for online updates to previously deployed networks. 

\par Likewise, in some emergencies such as UAV fails, the administrator can adjust the number of UAVs in the simulation and retrain the DRL algorithm accordingly. Since this process can be done in computation-sufficient conditions, the administrator may train the DRL algorithm for several versions with various numbers of UAVs in advance for backups. Moreover, some redundancy fault-tolerance mechanisms would also be beneficial to handle this emergency. For instance, the administrator could set up some backup UAVs, and these UAVs can be deployed to replace the faulty ones with minimal delay when a UAV breaks down. This mechanism may allow for real-time adaptation and maintain the performance of the communication system without significant downtime.

\subsection{Computational Complexity of the Proposed EMOPPO-VLH}

\par The total complexity of the proposed EMOPPO-VLH is given by $O(G_{max} \cdot n \cdot n_{evo} \cdot (\sum_{l=1}^L m_{l-1} \cdot m_l + M))$, in which $n$ is the number of tasks, $n_{evo}$ is the number of evolutionary iterations, and $L$ and $m_l$ refer to the number of layers and units in the deep network, respectively. The detailed analyses can be found in Appendix A.2.

\section{Simulation Results}
\label{sec:simulation results}

\par In this section, we provide extensive simulation results to demonstrate the performance and advantages of the proposed EMOPPO-VLH in addressing the formulated MOP.

\subsection{Simulation Setups}

\par
\begin{figure}[t]
	\centerline{\includegraphics[width=3.2in]{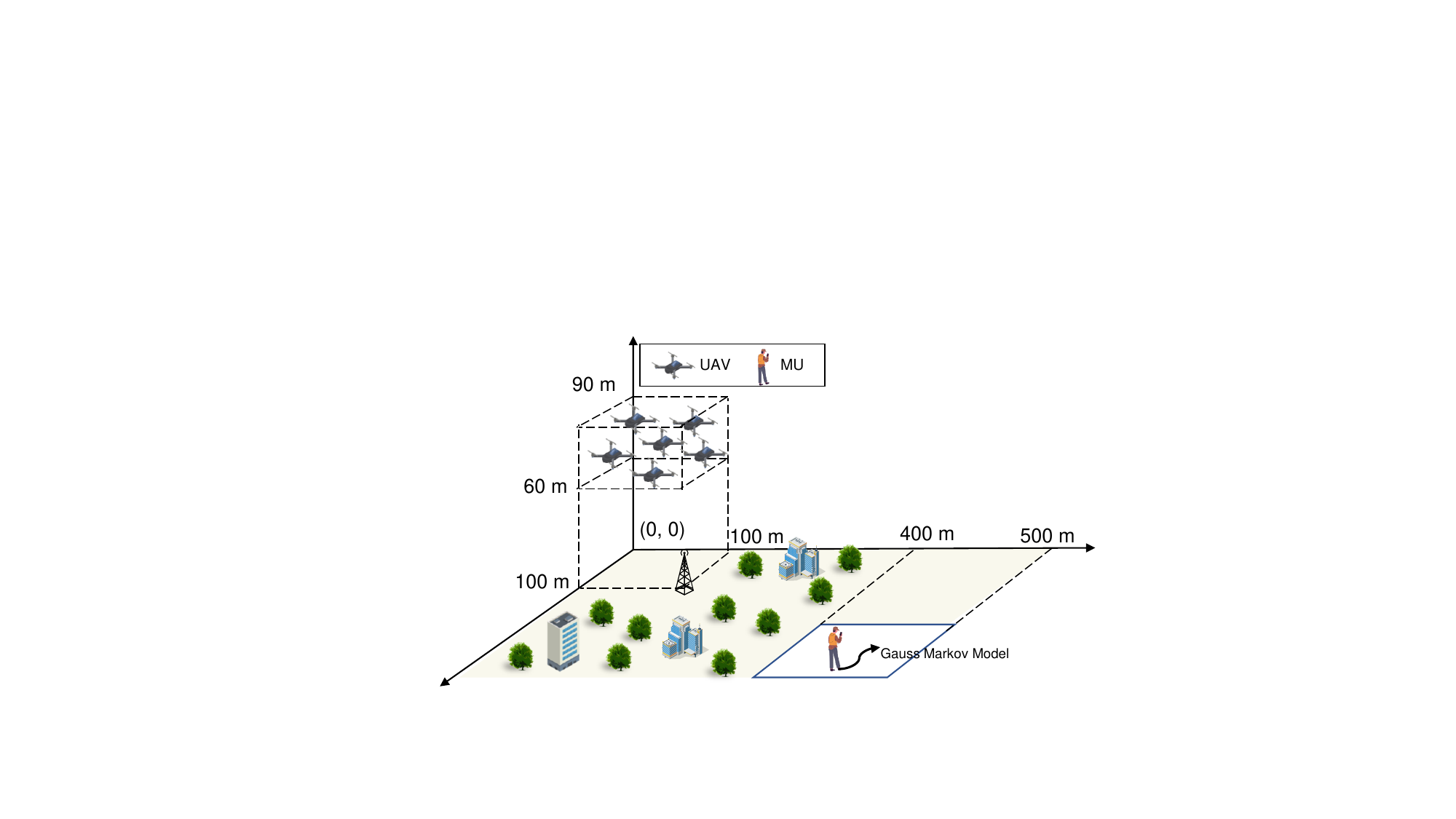}}
	\caption{The schematic map illustrates the simulation setup. UAVs are dispersed across a 100 m $\times$ 100 m region, and a terrestrial mobile user moves randomly within another 100 m $\times$ 100 m rectangular. The BS is positioned at coordinates (100, 100) in meters.}
	\label{fig:simulation}
\end{figure}

 We consider two different scales of UAV swarm, namely, small-scale and large-scale UAV swarms, containing 8 and 16 UAVs, respectively. The purpose of setting different scenario scales is to verify the scalability and robustness of our proposed algorithm under varying environmental sizes. By testing both small-scale and large-scale scenarios, we aim to demonstrate that our algorithm performs effectively across different operational contexts and can handle environments of varying sizes. Moreover, the UAVs are distributed in a rectangular area where each side measures $L_{max} = 100$ m. The mass of the UAVs ($m_{UAV}$), the collision distance ($d_{min}$), and the minimum and maximum altitudes ($H_{min}$ and $H_{max}$) are set as 2 kg, 0.5 m, 60 m, and 90 m, respectively. Additionally, the total noisy power spectral density, the path loss exponent, the total transmit power of each UAV, and the carrier frequency are set at -155 dBm/Hz, 2, 0.1 W, and 2.4 GHz, respectively. For other key parameters related to UAVs, we refer to \cite{sun2022secure}. The simulation considers a communication period of 300 seconds with each time slot lasting 1 second. Initially, the UAVs are randomly distributed within a rectangular area. In each time slot, the UAVs have maximum permissible horizontal ($d_{max}^h$) and vertical flight distances ($d_{max}^v$) of 20 m and 10 m, respectively. Furthermore, the terrestrial mobile user moves with an average speed of 1 m/s within a square area measuring 100 m $\times$ 100 m. For a more intuitive presentation, we provide a 2D diagram of the system in Fig.~\ref{fig:simulation}.

\par Moreover, the number of evenly distributed weight vectors $n$ is set to 15, with each vector corresponding to a specific learning task. For each learning task, both the policy and value networks consist of an LSTM layer with 128 neurons, followed by a three-layer fully connected neural network, each layer containing 256 neurons. Additionally, we use the tanh function as the activation function in each hidden layer. The parameters of the policy and value networks are updated through the Adam optimizer with a learning rate of 0.0001. The discount factor $\gamma = 0.99$ and the clip parameter $\epsilon = 0.2$ are also specified. Furthermore, the maximum evolution generations $G_{max}$, the task iterations of the warm-up stage $n_{warm}$, and the evolutionary stage $n_{evo}$ are set to 100, 60, and 10, respectively. The number of performance buffers is set to 50, with each buffer having a size of 2.

\subsection{Performance Indicators}

\par Different from single-objective optimization, the performance evaluation of multi-objective optimization problems is relatively complicated. In this work, we adopt two performance evaluation metrics to evaluate the proximity to the Pareto front, the diversity of the obtained policy of EMOPPO-VLH, including the inverted generational distance (IGD)~\cite{Cai2021AGrid-Based} and hypervolume (HV) \cite{Shang2021ASurvey}. The details of them are shown in Appendix B of the supplemental material.

\subsection{Baselines}

\par To evaluate the effectiveness of our proposed EMOPPO-VLH, we implement several types of comparison algorithms, including two multi-objective evolutionary algorithms (MOEAs), namely, a multi-objective evolutionary algorithm based on decomposition (MOEA/D) \cite{zhang2007moea} and multi-objective particle swarm optimization (MOPSO) \cite{Coello2002MOPSO}, two multi-policy MORLs, namely, evolutionary deep deterministic policy gradient (EDDPG) and evolutionary twin delayed DDPG (ETD3), standard evolutionary PPO (EPPO), and EPPO with gated recurrent unit (GRU). The baseline algorithms are described as follows:

\begin{itemize}
    
    \item MOEA/D: This method decomposes a multi-objective optimization problem into multiple scalar optimization sub-problems and optimizes them concurrently. The maximum number of evolutionary generations and the population size are both fixed at 100. Furthermore, the size of the neighborhood associated with each sub-problem is set to 10.

    \item MOPSO: This method extends particle swarm optimization to handle multi-objective problems, which uses Pareto dominance to guide the flight direction of the particles. Moreover, the number of generations, the population size, the repository size, and the division for the adaptive grid are set to 100, 50, 100, and 30, respectively.

    \item EDDPG: For performance evaluation, we design an EDDPG. DDPG is a well-known actor-critic reinforcement learning approach extensively employed in continuous control tasks. EDDPG is a modified version by replacing MOPPO with multi-objective DDPG, as outlined in Algorithm~\ref{alg:lstm-ppo}. It is noteworthy that the multi-objective DDPG is extended from the conventional single-policy DDPG.

    \item ETD3: To rigorously evaluate performance metrics, we develop another algorithm called ETD3. TD3 is an improved DDPG, which incorporates clipped double critic networks to ameliorate the overestimation of Q-values. Similar to EDDPG, ETD3 is derived by replacing MOPPO with multi-objective TD3. Note that multi-objective TD3 is extended from single-policy TD3.

    \item EPPO: The original EPPO is employed as a benchmark strategy to showcase the efficiency of our proposed enhancement measures.

    \item EPPO with GRU: We develop the EPPO with GRU as a benchmark strategy to showcase the effectiveness of the introduced LSTM network structure.

\end{itemize}

\par In MOEA/D and MOPSO, each solution represents the UAV trajectory control and excitation current weights across all time slots. For a fair comparison, EMOPPO-VLH, EMOPPO, EDDPG, and ETD3 are configured with identical parameters. Furthermore, optimization objective 1, as presented in Eq.~(\ref{eq:objective_1}), is deemed paramount in the majority of scenarios. Consequently, from the Pareto set, we opt for the solution exhibiting optimal performance in Objective 1 as the final solution for all considered algorithms.

\par Note that the computational complexities of various comparison algorithms, including MOEA/D and MOPSO, differ from those of the proposed method. Specifically, the computational complexity of MOEA/D and MOPSO is dominated by operations such as population initialization, neighborhood structure construction, and population updating, leading to an overall complexity of $O(G \times N^2)$, where $G$ represents the number of generations and $N$ is the population size. On the other hand, the computational complexity of EDDPG, ETD3, EPPO, and EPPO with GRU are similar, which are $O(G_{max} \cdot n \cdot n_{evo} \cdot (\sum_{l=1}^L m_{l-1} \cdot m_l + M))$, in which $n$ is the number of tasks, $n_{evo}$ is the number of evolutionary iterations, and $L$ and $m_l$ refer to the number of layers and units in the deep network, respectively. Compared to traditional evolutionary algorithms like MOEA/D and MOPSO, the proposed method leverages deep neural networks to handle high-dimensional state and action spaces more efficiently. Although the asymptotic complexities might appear similar, the proposed method can be more scalable and effective in practice due to the generalization capabilities of neural networks.

\subsection{Performance Evaluation}

\par
\begin{figure}[t] 
    \centering  
    \includegraphics[width=0.45\textwidth]{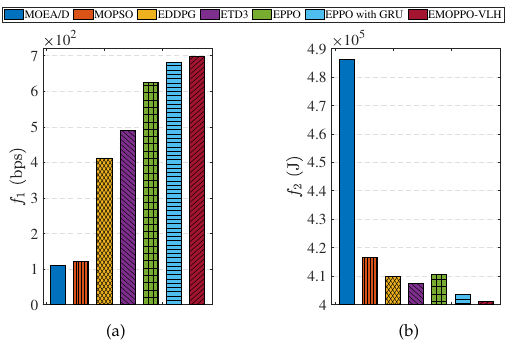}  
    \caption{Optimization results obtained by various algorithms in the small-scale scenario. (a) $f_1$ obtained by different algorithms. (b) $f_2$ obtained by different algorithms.}  
    \label{zhuxingtu_8}
\end{figure}

\begin{figure}[t] 
    \centering  
    \includegraphics[width=0.45\textwidth]{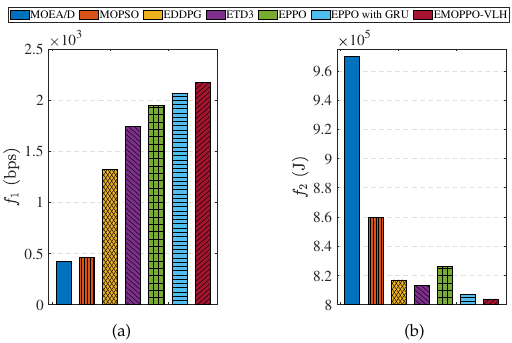}  
    \caption{Optimization results obtained by various algorithms in the large-scale scenario. (a) $f_1$ obtained by different algorithms. (b) $f_2$ obtained by different algorithms.}  
    \label{zhuxingtu_16}
\end{figure}

 Figs. \ref{zhuxingtu_8} and~\ref{zhuxingtu_16} display the numerical optimization results obtained by the aforementioned methods, including the total achievable rate ($f_1$) and the total energy consumption of the UAVs ($f_2$) in the small-scale and large-scale scenarios, respectively. As can be observed, the proposed EMOPPO-VLH outperforms all other optimization approaches across both scales. The higher total achievable rate obtained by EMOPPO-VLH indicates that the transmission process is more resistant to interference and will be completed in less time, which can further reduce the energy consumption of the UAVs. As such, the high performance of EMOPPO-VLH indicates its suitability over other algorithms for solving the formulated MOP.



\par
\begin{figure}[t]
    \centering
    \subfigure[Small-scale scenario]{
        \begin{minipage}[t]{0.45\textwidth}  
            \centering  
            \includegraphics[width=0.9 \linewidth]{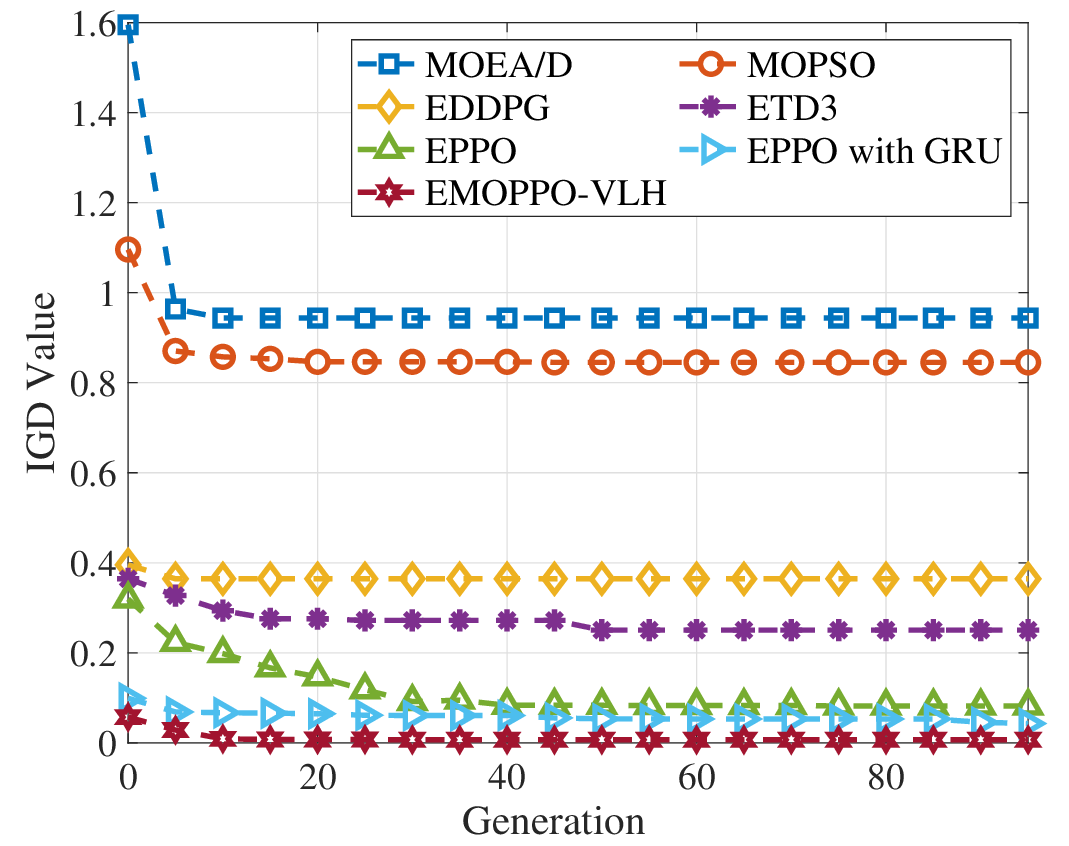}
        \end{minipage}
    }\\  
    \subfigure[Large-scale scenario]{
        \begin{minipage}[t]{0.45\textwidth}  
            \centering  
            \includegraphics[width=0.9 \linewidth]{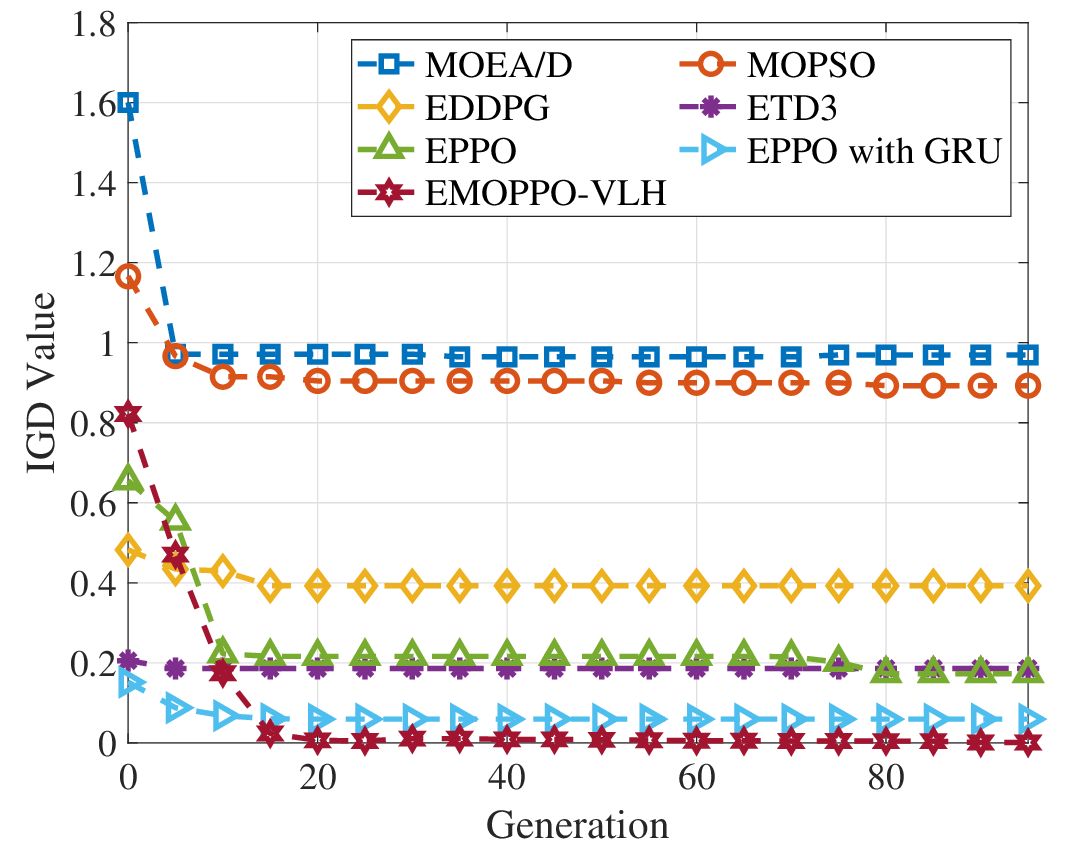}
        \end{minipage}
    }
    \caption{Comparison of convergences of the proposed algorithms improved by LSTM and GRU, as well as the other baseline algorithms based on IGD metrics.}  
    \label{fig:IGD}
\end{figure}

\begin{figure}[t]
    \centering
    \subfigure[Small-scale scenario]{
        \begin{minipage}[t]{0.45\textwidth}  
            \centering  
            \includegraphics[width=0.9 \linewidth]{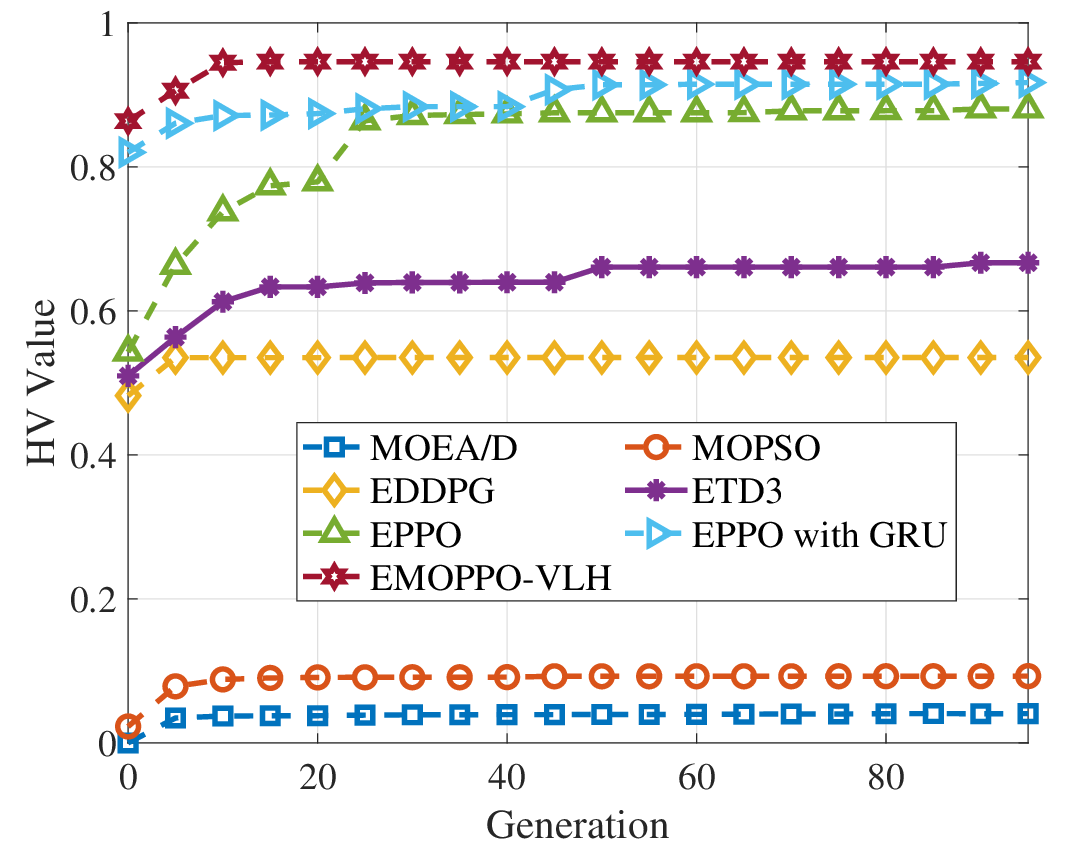}
        \end{minipage}
    }\\  
    \subfigure[Large-scale scenario]{
        \begin{minipage}[t]{0.45\textwidth}  
            \centering  
            \includegraphics[width=0.9 \linewidth]{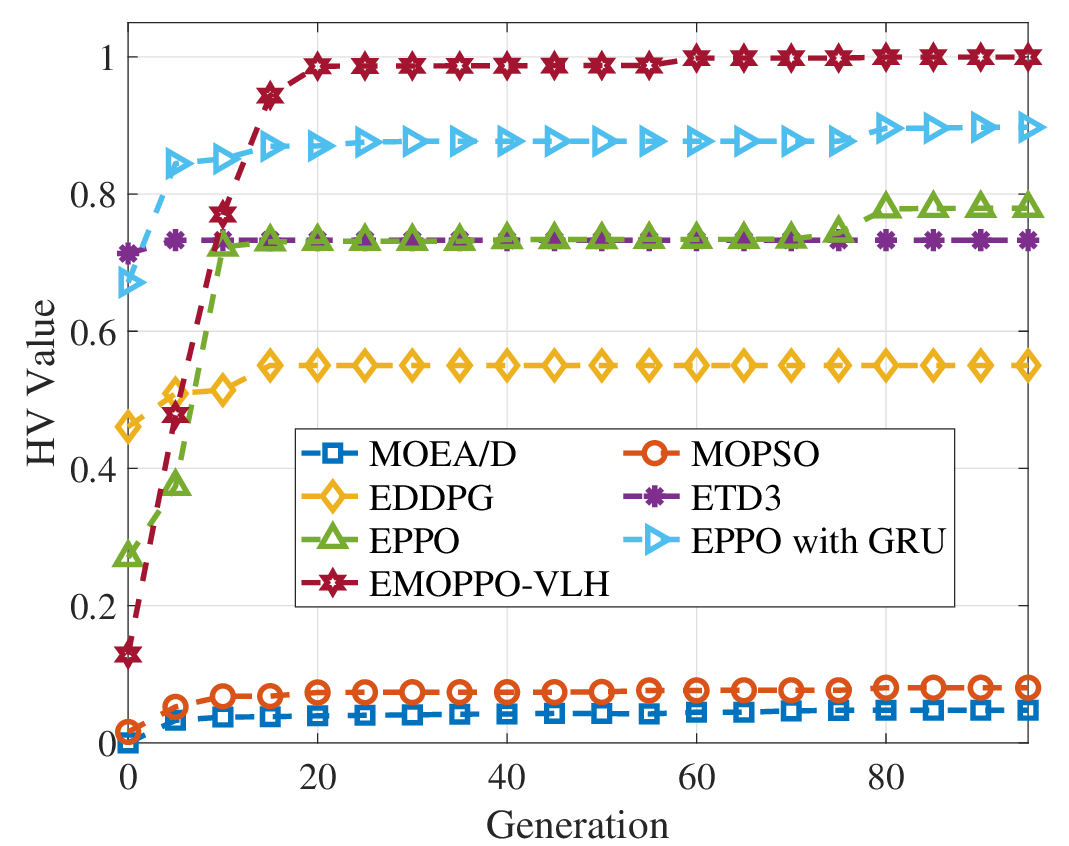}
        \end{minipage}
    }
    \caption{Comparison of convergences of the proposed algorithms improved by LSTM and GRU, as well as the other baseline algorithms based on IGD metrics.}  
    \label{fig:HV}
\end{figure}

 To facilitate demonstration, the outcomes for the IGD and HV metrics in smaller and larger scale networks are shown in Figs.~\ref{fig:IGD} and \ref{fig:HV}, respectively. A clear observation is that both MOEA/D and MOPSO underperform other algorithms, which fail to achieve a desirable Pareto front across all scenarios. This underperformance can be attributed to the inherent challenges of managing high-dimensional MOP in dynamic settings. Notably, MOEAs often invest significant computational effort in generating non-dominated policies, making it difficult to converge in a short time. More precisely, MOEAs need to find non-dominated policies characterized by extensive decision variables (\emph{i.e.}, $4 \cdot N \cdot T$), which is time-consuming. In addition, the highly dynamic and uncertain nature of the considered UAV-assisted communication environment frequently triggers the execution of MOEAs, thereby leading to considerable computational overhead and slow response. This explains the observed underperformance of MOEA/D and MOPSO in addressing the formulated MOP as presented in this work.

\par We observe that the proposed EMOPPO-VLH can achieve the smallest IGD values and the largest HV values across all scenarios, which can demonstrate its superiority over the other five algorithms. The reason is that LSTM-based MOPPO can capture time dependence, thereby enhancing the quality of each policy during the process of iterations. As such, the hyper-sphere-based task selection approach gives the obtained non-dominated policies wider coverage, representing more trade-offs that can meet different requirements. Moreover, the EMOPPO-VLH with LSTM is still better than EMOPPO-GRU, especially in the large-scale scenario. The reason may be that GRU has a simpler structure that has fewer gates than LSTM, which offers computational efficiency but slightly decreases the ability to capture the dynamic and complex temporal correlations in the formulated problem. As such, the LSTM may be more suitable for improving the ability to capture temporal correlations of the proposed algorithm. Such simulation results also validate the effectiveness of the proposed enhancement measures of EMOPPO-VLH.

\subsection{The Impacts of System Parameters}

\par In this subsection, we evaluate the impacts of the system parameters, including the parameters in the reward function, user mobility, and UAV numbers. Additionally, the robustness of the proposed approach is assessed under various unexpected scenarios, such as imperfect phase synchronization, a damaged UAV component, and positional jitters of the UAVs. Detailed results and analyses are provided in the Appendix C of the supplemental material.

%
%
\section{Discussion}
\label{sec:discussion}

\par In this section, we discuss the effectiveness, reasonableness, and scalability of the proposed method, and the details are shown in Appendix D of the supplemental material.

\section{Conclusion}
\label{sec:conclusion}

\par In this work, we investigated aerial reliable communications for terrestrial mobile user in UAV networks by using CB. First, we considered a typical scenario where a UAV swarm transmits the collected data to remote terrestrial mobile users by using CB, contending with interference from non-associated BSs and time-varying channels. Additionally, we introduced a more realistic random walk model to simulate the movement of the user. Subsequently, we formulated an MOP with the dual goals of maximizing the achievable rate and minimizing the energy consumption for the UAVs. Given the NP-hard nature of the formulated MOP and the highly dynamic environment, we proposed an EMOPPO-VLH to tackle this problem. The algorithm can obtain a set of non-dominated policies with various trade-offs, thereby meeting different user preferences. Simulation results illustrated that the proposed algorithm outperforms other benchmark algorithms in both small-scale and large-scale scenarios. The results for IGD and HV confirm that the proposed EMOPPO-VLH exhibits superior convergence and diversity. Additional simulations demonstrate the scalability and robustness of the proposed CB-based method under different system parameters and various unexpected circumstances. In the future, we will explore UAV-enabled CB involving different types of UAVs.

\bibliographystyle{IEEEtran}
\bibliography{myref.bib}

\vspace{-10 mm}
\begin{IEEEbiography}[{\includegraphics[width=1in,height=1.25in,clip,keepaspectratio]{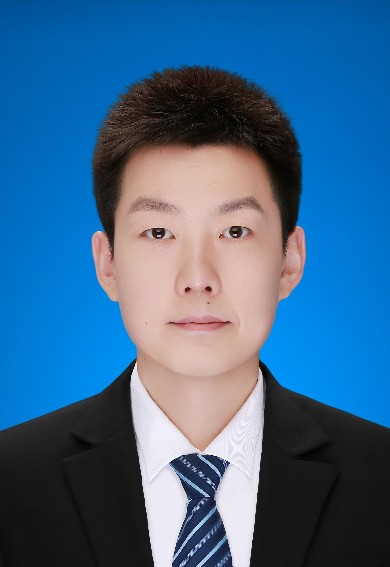}}]{Geng Sun}
(S'17-M'19) received the B.S. degree in communication engineering from Dalian Polytechnic University, and the Ph.D. degree in computer science and technology from Jilin University, in 2011 and 2018, respectively. He was a Visiting Researcher with the School of Electrical and Computer Engineering, Georgia Institute of Technology, USA. He is a Professor in College of Computer Science and Technology at Jilin University, and his research interests include wireless networks, UAV communications, collaborative beamforming and optimizations.
\end{IEEEbiography}

\vspace{-10 mm}
\begin{IEEEbiography}[{\includegraphics[width=1in,height=1.25in,clip,keepaspectratio]{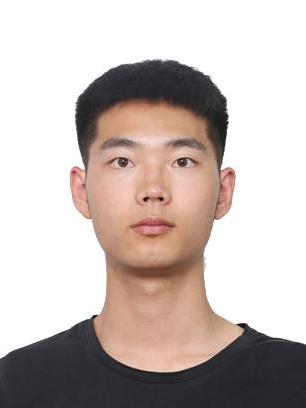}}]{Jian Xiao}
received the B.S. degree in computer science and technology from Harbin University of Science and Technology in 2022. He is currently working towards the M.S. degree at the College of Computer Science and Technology, Jilin University. His research interests include UAV networks, antenna arrays, and optimization.
\end{IEEEbiography}

\vspace{-10 mm}
\begin{IEEEbiography}[{\includegraphics[width=1in,height=1.25in,clip,keepaspectratio]{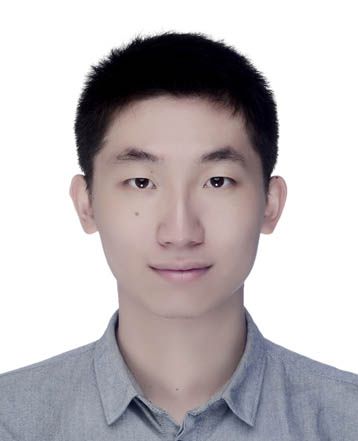}}]{Jiahui Li} received his B.S. in Software Engineering, and M.S. and Ph.D. in Computer Science and Technology from Jilin University, Changchun, China, in 2018, 2021, and 2024, respectively. He was a visiting PhD student at the Singapore University of Technology and Design (SUTD). He currently serves as an assistant researcher in the College of Computer Science and Technology at Jilin University. His current research focuses on integrated air-ground networks, UAV networks, wireless energy transfer, and optimization.
\end{IEEEbiography}


\vspace{-10 mm}
\begin{IEEEbiography}[{\includegraphics[width=1in,height=1.25in,clip,keepaspectratio]{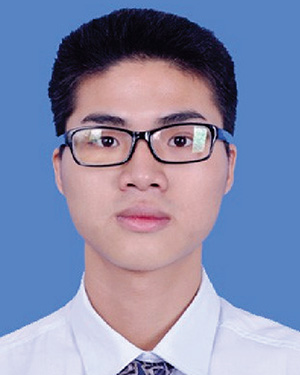}}]{Jiawen Kang}
received the Ph.D. degree from the Guangdong University of Technology, China in 2018. From 2018 to 2021, he was a Post-Doctoral Fellow with Nanyang Technological University, Singapore. He is currently a Full Professor with the Guangdong University of Technology. His research interests include blockchain, security, and privacy protection in wireless communications and networking.
\end{IEEEbiography}

\vspace{-10 mm}
\begin{IEEEbiography}[{\includegraphics[width=1in,height=1.25in,clip,keepaspectratio]{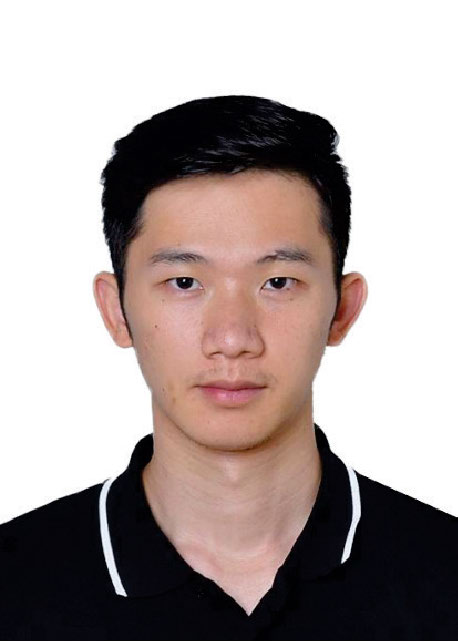}}]{Jiacheng Wang}
received the Ph.D. degree from the School of Communication and Information Engineering, Chongqing University of Posts and Telecommunications, Chongqing, China. He is cur-rently a Research Associate in computer science and engineering with Nanyang Technological University, Singapore. His research interests include wireless sensing and semantic communications.
\end{IEEEbiography}

\vspace{-10 mm}
\begin{IEEEbiography}[{\includegraphics[width=1in,height=1.25in,clip,keepaspectratio]{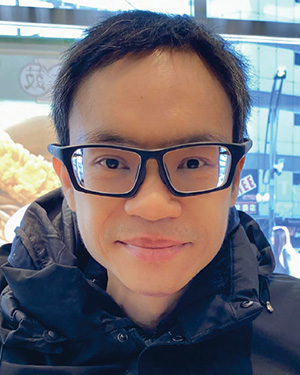}}]{Dusit Niyato}
(Fellow, IEEE) received the B.Eng. degree from the King Mongkuts Institute of Tech-nology Ladkrabang (KMITL), Thailand, in 1999, and the Ph.D. degree in electrical and computer engineering from the University of Manitoba, Canada, in 2008. He is currently a Professor with the School of Computer Science and Engineer-ing, Nanyang Technological University, Singa-pore. His research interests include the Internet of Things (IoT), machine learning, and incentive mechanism design.
\end{IEEEbiography}

\vspace{-10 mm}
\begin{IEEEbiography}[{\includegraphics[width=1in,height=1.25in,clip,keepaspectratio]{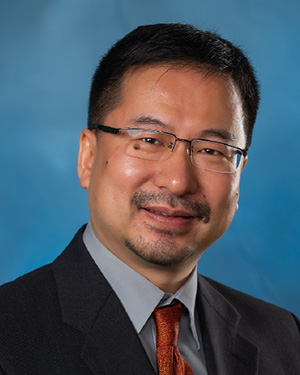}}]{Shiwen Mao}
(Fellow, IEEE) is a Professor and the Earle C. Williams Eminent Scholar Chair, and the Director of the Wireless Engineering Research and Education Center, Auburn University, Auburn, AL, USA. His research interest includes wireless networks, multimedia communications, and smart grid. He received the IEEE ComSoc MMTC Outstanding Researcher Award in 2023, the IEEE ComSoc TC-CSR Distinguished Technical Achievement Award in 2019, and the NSF CAREER Award in 2010. He is a co-recipient of the 2022 Best Journal Paper Award of IEEE ComSoc eHealth Technical Committee, the 2021 Best Paper Award of Elsevier/Digital Communications and Networks (KeAi), the 2021 IEEE Internet of Things Journal Best Paper Award, the 2021 IEEE Communications Society Outstanding Paper Award, the IEEE Vehicular Technology Society 2020 Jack Neubauer Memorial Award, the 2018 ComSoc MMTC Best Journal Paper Award and the 2017 Best Conference Paper Award, the 2004 IEEE Communications Society Leonard G. Abraham Prize in the Field of Communications Systems, and several ComSoc technical committee and conference best paper/demo awards. He is the Editor-in-Chief of IEEE TRANSACTIONS ON COGNITIVE COMMUNICATIONS AND NETWORKING. He is a Distinguished Lecturer of IEEE Communications Society and the IEEE Council of RFID.
\end{IEEEbiography}


\end{document}